\documentclass[12pt]{amsart}
\newcommand{\beginsupplement}{%
        \setcounter{table}{0}
        \renewcommand{\thetable}{S\arabic{table}}%
        \setcounter{figure}{0}
        \renewcommand{\thefigure}{S\arabic{figure}}%
        \setcounter{section}{0}
        \renewcommand{\thesection}{S\arabic{section}}%
     }

\usepackage{upgreek}
\usepackage{natbib}
\usepackage[table]{xcolor}
\usepackage{amsaddr}
\usepackage{pifont}
\usepackage{mathtools}
\usepackage{amssymb, times,caption}
\usepackage{wrapfig}
\usepackage{bm}
\usepackage[scr]{rsfso}
\usepackage{mathrsfs}
\usepackage[T1]{fontenc}
\usepackage{hyphenat}
\usepackage{xcolor}
\usepackage{tikz}
\usetikzlibrary{shapes}
\usepackage{float,amsmath}
\usepackage{sidecap}
\usepackage{epsfig}
\usepackage{eucal}
\usepackage{graphicx} 
\usepackage{subcaption}
\usepackage[T1]{fontenc}
\usepackage{mathabx}

\newtheorem{statement}{Statement}
\newtheorem{appendix-statement}{Appendix Statement}
\newtheorem{appendix-corollary}{Appendix Corollary}

\newtheorem*{control-problem}{Control Problem}

\def\Z{\CMcal{Z}}
\def\S{\CMcal{S}}
\def\F{\CMcal{F}}
\definecolor{background-color}{gray}{0.98}

\usepackage{float,amsmath}
\usepackage{epsfig}
\usepackage{eucal}
\usepackage{graphicx} 
\usepackage[T1]{fontenc}
\usepackage{mathabx}

\newtheorem{lemma}{Lemma}  
\def\squareforqed{\hbox{\rlap{$\sqcap$}$\sqcup$}}
\def\qed{\ifmmode\squareforqed\else{\unskip\nobreak\hfil
\penalty50\hskip1em\null\nobreak\hfil\squareforqed
\parfillskip=0pt\finalhyphendemerits=0\endgraf}\fi}

\newcommand{\argmax}{{\mathrm{argmax}}}

\newcommand{\ignore}[1]{}

\newcommand{\R}{\mathbb{R}}

\begin{document}
\title{Hidden Markov Modeling for Maximum Probability Neuron Reconstruction}
\author[T. L. Athey]{Thomas L. Athey}
\address{Center for Imaging Science \& Kavli Neuroscience Discovery Institute \\ \&
    Department of Biomedical Engineering\\ The Johns Hopkins
    University}
\email[T. L. Athey]{tathey1@jhu.edu}
\author[D. J. Tward]{Daniel J. Tward}
\address{Department of Computational Medicine \& Department of Neurology\\ University of California at Los Angeles}
\email[D. Tward]{DTward@mednet.ucla.edu}
\author[U. Mueller]{Ulrich Mueller}
\address{Department of Neuroscience\\ The Johns Hopkins
    University}
\email[U. Mueller]{umuelle3@jhmi.edu}
\author[J. T. Vogelstein]{Joshua T. Vogelstein}
\address{Center for Imaging Science \& Kavli Neuroscience Discovery Institute  \\ \&
    Department of Biomedical Engineering\\ The Johns Hopkins
    University\\ 301 Clark Hall\\ Baltimore, MD 21218}
\email[J. T. Vogelstein]{jovo@jhu.edu}
\author[M. I. Miller]{Michael I. Miller}
\address{Center for Imaging Science \& Kavli Neuroscience Discovery Institute  \\ \&
    Department of Biomedical Engineering\\ The Johns Hopkins
    University\\ 301 Clark Hall\\ Baltimore, MD 21218}
\email[M. I. Miller]{mim@cis.jhu.edu}
\date{\today}

\begin{abstract} 
Recent advances in brain clearing and imaging have made it possible to image entire mammalian brains at sub-micron resolution. These images offer the potential to assemble brain-wide atlases of neuron morphology, but manual neuron reconstruction remains a bottleneck. Several automatic reconstruction algorithms exist, but most focus on single neuron images. In this paper, we present a probabilistic reconstruction method, ViterBrain, which combines a hidden Markov state process that encodes neuron geometry with a random field appearance model of neuron fluorescence. Our method utilizes dynamic programming to compute the global maximizers of what we call the ``most probable'' neuron path. Our most probable estimation method models the task of reconstructing neuronal processes in the presence of other neurons, and thus is applicable in images with several neurons. Our method operates on image segmentations in order to leverage cutting edge computer vision technology. We applied our algorithm to imperfect image segmentations where false negatives severed neuronal processes, and showed that it can follow axons in the presence of noise or nearby neurons. Additionally, it creates a framework where users can intervene to, for example, fit start and endpoints. The code used in this work is available in our open-source Python package \texttt{brainlit}.
\end{abstract}

\maketitle
\section{Introduction}

Neuron morphology has been a central topic in neuroscience for over a century, as it is the substrate for neural connectivity, and serves as a useful basis for neuron classification. Technological advances in brain clearing and imaging have allowed scientists to probe neurons that extend throughout the brain, and branch hundreds of times \citep{winnubst2019reconstruction}. It is becoming feasible to assemble a brainwide atlas of cell types in the mammalian brain which would serve as a foundation to understanding how the brain operates as an integrated circuit, or how it fails in neurological disease. One of the main bottlenecks in assembling such an atlas is the manual labor involved in neuron reconstruction.

In an effort to accelerate reconstruction, many automated reconstruction algorithms have been proposed, especially over the last decade. In 2010, the DIADEM project brought multiple institutions together to consolidate existing algorithms, and stimulate further progress by generating open access image datasets, and organizing a contest for reconstruction algorithms \citep{peng2015diadem}. Several years later, the BigNeuron project continued the legacy of DIADEM, this time establishing a common software platform, Vaa3D, on which many of the state of the art algorithms were implemented \citep{peng2010v3d}. \citet{acciai2016automated} offers a review of notable reconstruction algorithms up to, and through, the BigNeuron project.

Previous approaches to automated neuron reconstruction have used shortest path/geodesic computation \citep{peng2010automatic, Wang-Roysam-2011, turetken2013reconstructing}, minimum spanning trees \citep{yang2019fmst}, Bayesian estimation \citep{radojevic-2017}, tracking \citep{choromanska2012automatic, dai2019deep}, and deep learning \citep{friedmann2020mapping, zhou2018deepneuron, li2017deep}. Methods have also been developed to enhance, or extend existing reconstruction algorithms \citep{wang2019teravr, wang2017ensemble, peng2017automatic, chen2015smarttracing}. Also, some works focus on the subproblem of resolving different neuronal processes that pass by closely to each other \citep{APP2, Li2019PreciseSO, quan2016neurogps}.

Both DIADEM and BigNeuron initiatives focused on the task of single neuron reconstruction so most associated algorithms fail when applied to images with several neurons. However, robust reconstruction in multiple neuron images is essential in order to assemble brainwide atlases of neuron morphology.

We propose a probabilistic model-based algorithm, ViterBrain, that operates on imperfect image segmentations to efficiently reconstruct neuronal processes. Our estimation method does not assume that the image outside the reconstruction is background, and thus allows for the existence of other neurons. Our approach draws upon two major subfields in Computer Vision, appearance modeling and hidden Markov models, and generates globally optimal solutions using dynamic programming. The states of our model are locally connected segments. We score the state transitions using appearance models such as exhibited by Kass and Cohen's early works \citep{kass1988snakes, COHEN1991211} on active shape modeling and their subsequent application by Wang et al. \citep{Wang-Roysam-2011} for neuron reconstruction. For our own approach we exploit foreground-background models of image intensity for the data likelihood term in the hidden-Markov structure. We quantify the image data using KL-divergence and intensity autocorrelation in order to validate our model assumptions.

Our probabilistic models are hidden Markov random fields, but we reduce the computational structure to a hidden Markov model (HMM) since the latent axonal structures have an absolute ordering. 
Hidden Markov modeling (HMM) involves two sequences of variables, one is observed and one is hidden. A popular application of HMM's is in speech recognition where the observed sequence is an audio signal, and the hidden variables a sequence of words \citep{rabiner1986introduction}. In our setting, the observed data is the image, and the hidden data is the contour representation of the axon or dendrite's path.

The key advantages to using HMMs in this context are, first, that neuronal geometry can be explicitly encoded in the state transition distribution. We utilize the Frenet representation of curvature in our transition distribution, which we have studied previously in \citet{athey2021spline,Khaneja98dynamicprogramming}. 
Secondly, globally optimal estimates can be computed efficiently using dynamic programming in HMMs \citep{forney1973viterbi, Khaneja98dynamicprogramming}. The well-known Viterbi algorithm computes the MAP estimate of the hidden sequence in an HMM. Our approach, inspired by the Viterbi algorithm, also computes globally optimal estimates. Thus, our optimization method is not susceptible to local optima that exist in filtering methods, or gradient methods in active shape modeling. 

\begin{figure}[ht]
    \centering
    \includegraphics[width=\textwidth]{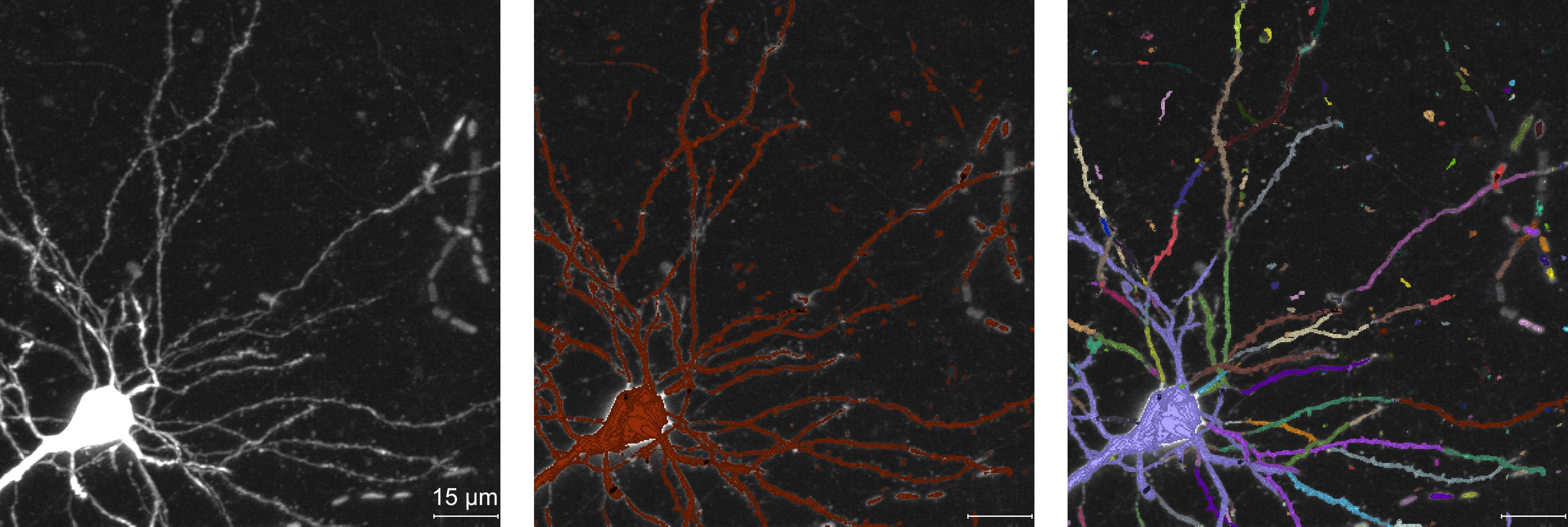}
    \caption{Left panel shows an image subvolume from the MouseLight project containing a single neuron. The middle panel shows the same image overlaid with a binary image mask in brown. This mask was generated by the random forest based software Ilastik \citep{berg2019ilastik} and illustrates the typical output of an image segmentation model. The right panels shows the same binary image mask, with a different color for each connected component. The variety of colors shows that the neuron has been severed into several pieces. All panels are maximum intensity projections (MIPs).}
    \label{fig:components}
\end{figure}

In this work, we apply our hidden Markov modeling framework to the output of low-level image segmentation models. Convolutional neural networks have shown impressive results in image segmentation \citep{ronneberger2015u}, but it only takes a few false negatives to sever neuronal processes that are often as thin as one micron (Figure \ref{fig:components}). Our method strings together the locally connected components of the binary image masks into a reconstruction with a global ordering. Thus, our method is modular enough to leverage state of the art methods in machine learning for image segmentation.

We apply our method to data from the MouseLight project at Janelia Research Campus \citep{winnubst2019reconstruction}, and focus on the endpoint control problem in a single neuronal process i.e. start and end points are fixed. We introduce the use of Frechet distance to quantify the precision of reconstructions and show that our method has comparable precision to state of the art, when the algorithms are successful.

\section{Results}

\subsection{Overview of ViterBrain}

\begin{figure}[ht]
    \centering
    \includegraphics[width=\textwidth]{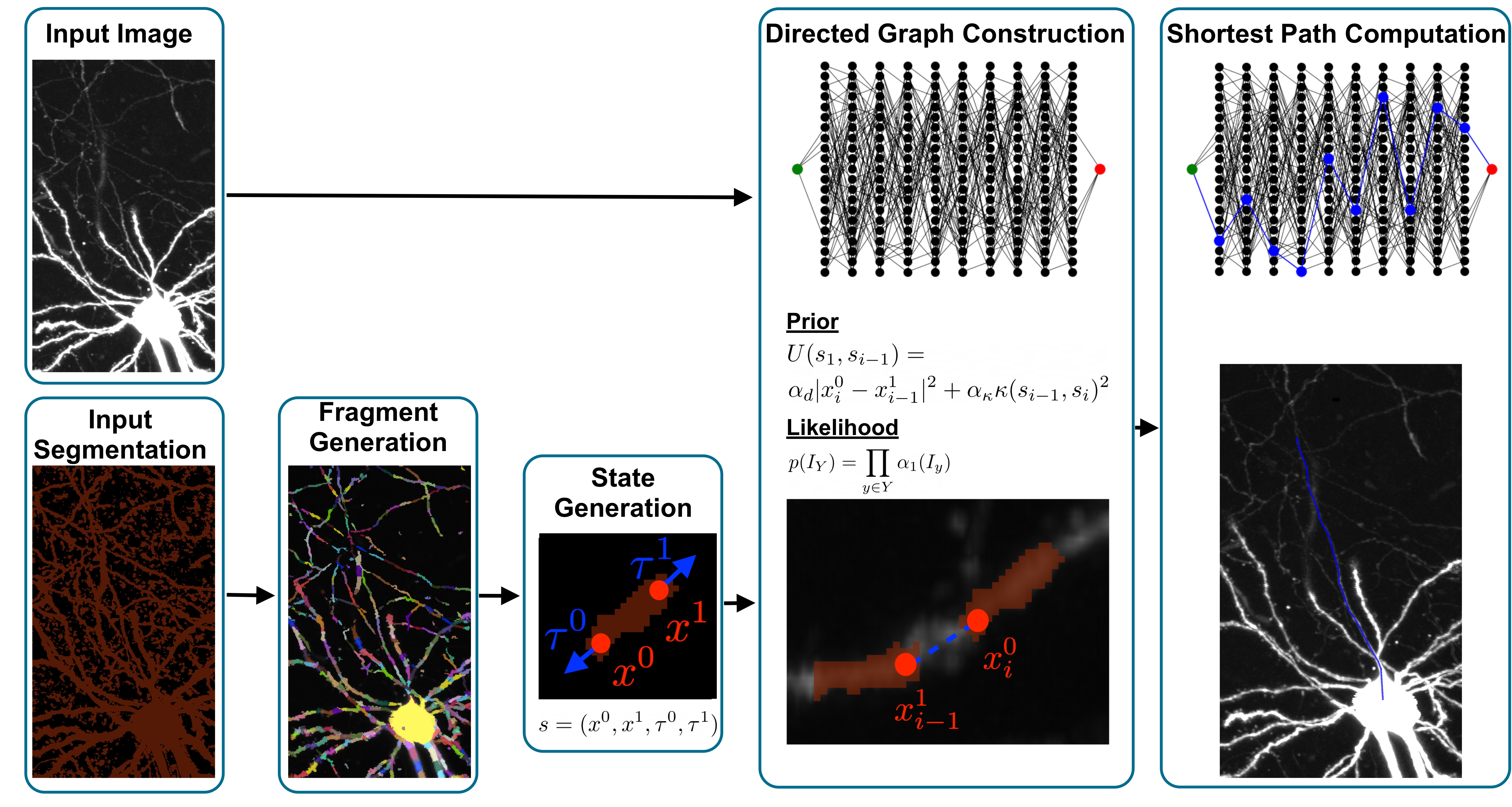}
    \caption{Summary of our algorithm. The algorithm takes in an image and a binary mask that might have severed, or fused neuronal processes. First, the mask is processed into a set of fragments. For each fragment, the endpoints and endpoint orientations are estimated and added to the state space. Next, transition probabilities are computed from both the image and state data to generate a directed graph reminiscent of the trellis graph in classic hidden Markov modeling. Finally, a shortest path algorithm is applied to compute the maximally probable state sequence connecting the start to the end state.}
    \label{fig:algorithm}
\end{figure}

Viterbrain takes in an image, and associated neuron mask produced by some image segmentation model. The algorithm starts by processing the mask into neuron \textit{fragments} and estimating fragment endpoints and orientation to generate the \textit{states}. Next, both the prior and likelihood terms of the transition probabilities are computed to construct a directed graph reminiscent of the the trellis from the HMM work in \citet{forney1973viterbi}. Lastly, the shortest path between states is computed with dynamic programming. We have proven that the shortest path is the ``most probable'' estimate state sequence formed by a neuronal process (Statement \ref{maximum-likely-state-sequences}). %Our most probable estimation procedure is different from maximum likelihood or maximum a posteriori estimation because it only considers the observed data located at the given state sequence. Thus, it does not assume that observed data outside of the state sequence is background, allowing for the possibility of other neurons. 
An overview of our algorithm shown in Figure \ref{fig:algorithm} and details are given in the Methods section. We validated our algorithm on subvolumes of one of the MouseLight whole-brain images \citep{winnubst2019reconstruction}. The image was acquired via serial two-photon tomography at a resolution of $0.3 \mu m \times 0.3 \mu m \times 1 \upmu$m per voxel. Viterbrain is available in our open source Python package \texttt{brainlit}: \texttt{http://brainlit.neurodata.io/}.

\subsection{Modeling Image Intensity}

\begin{figure}[ht]
    \centering
    \includegraphics[width=\textwidth]{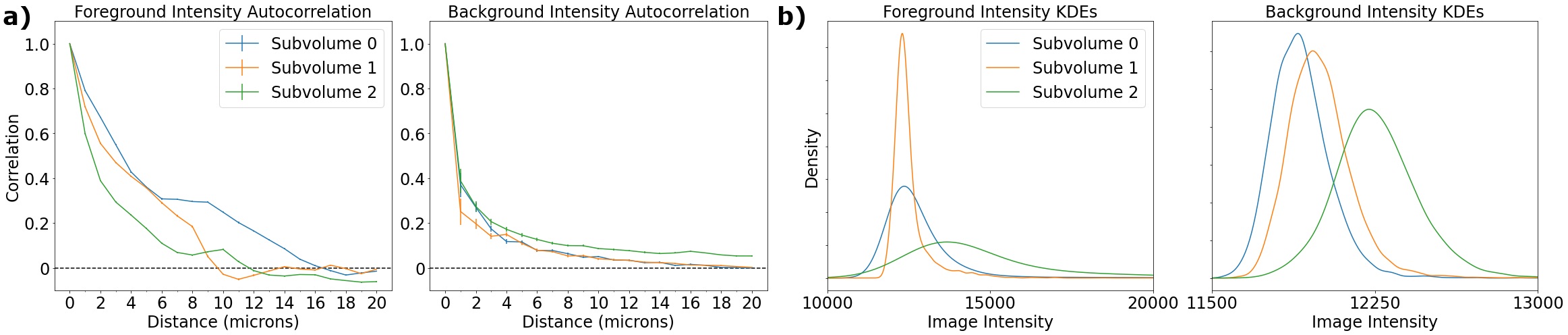}
    \caption{Characterization of voxel intensity distributions in three different subvolumes of one of the Mouselight whole-brain images. a) Correlation of intensities between voxels at varying distances from each other. The curves show that intensities are only weakly correlated ($\rho<0.4$) at a distance of $>10$ microns for foreground voxels, or $>2$ for background voxels. Error bars represent a single standard deviation of the Fisher z-transformation of the correlation coefficient. Each curve was generated from all pairs of 5000 randomly sampled voxels. b) Kernel density estimates (KDEs) of foreground and background intensity distributions. A subset of the voxels in each subvolume was manually labeled, then used to train an Ilastik model to classify the remaining voxels. Each KDE was generated from $5000$ voxels, according to the Ilastik classifications. KDEs were computed using scipy's Gaussian KDE function with default parameters \citep{2020SciPy-NMeth}.}
    \label{fig:fgbg}
\end{figure}

Figure \ref{fig:fgbg}a shows the correlations of image intensities between voxels at varying distances of separation. As is typical for natural images, voxels that are close by each other have positively correlated intensities, and those farther away are uncorrelated. In the case of foreground voxels, correlations become weak beyond a distance of about $10$ microns, with background voxel correlation decaying rapidly. This lends support to our assumption that voxel intensities are conditionally independent processes, conditioned on the foreground/background model (Eq. \ref{eq:alpha-product-probability}). This assumption is one of the central features of our model because it provides for computational tractability.

Figure \ref{fig:fgbg}b shows kernel density estimates (KDEs) of the foreground and background image intensity distributions. The distributions vary greatly between the three image subvolumes, implying that modeling the image process as homogeneous throughout the whole brain would be inappropriate. Additionally, the distributions do not appear to be either Gaussian or Poisson. Indeed, Kolmogorov-Smirnov tests rejected the null hypothesis for both Gaussian and Poisson goodness of fit in all cases, with all p-values below $10^{-16}$. For that reason, we exploit the independent increments properties of Poisson emission conditioned on the underlying intensity model, but do not assume that the marginal probabilities are Poisson (or Gaussian), instead, we estimate the intensity distributions from the data itself using KDEs (denoted $\alpha_0(\cdot),\alpha_1(\cdot)$ in Section \ref{sec:appearance}).

\subsection{Maximally Probable Axon Reconstructions}
\begin{figure}[ht]
    \centering
    \includegraphics[width=0.6\textwidth]{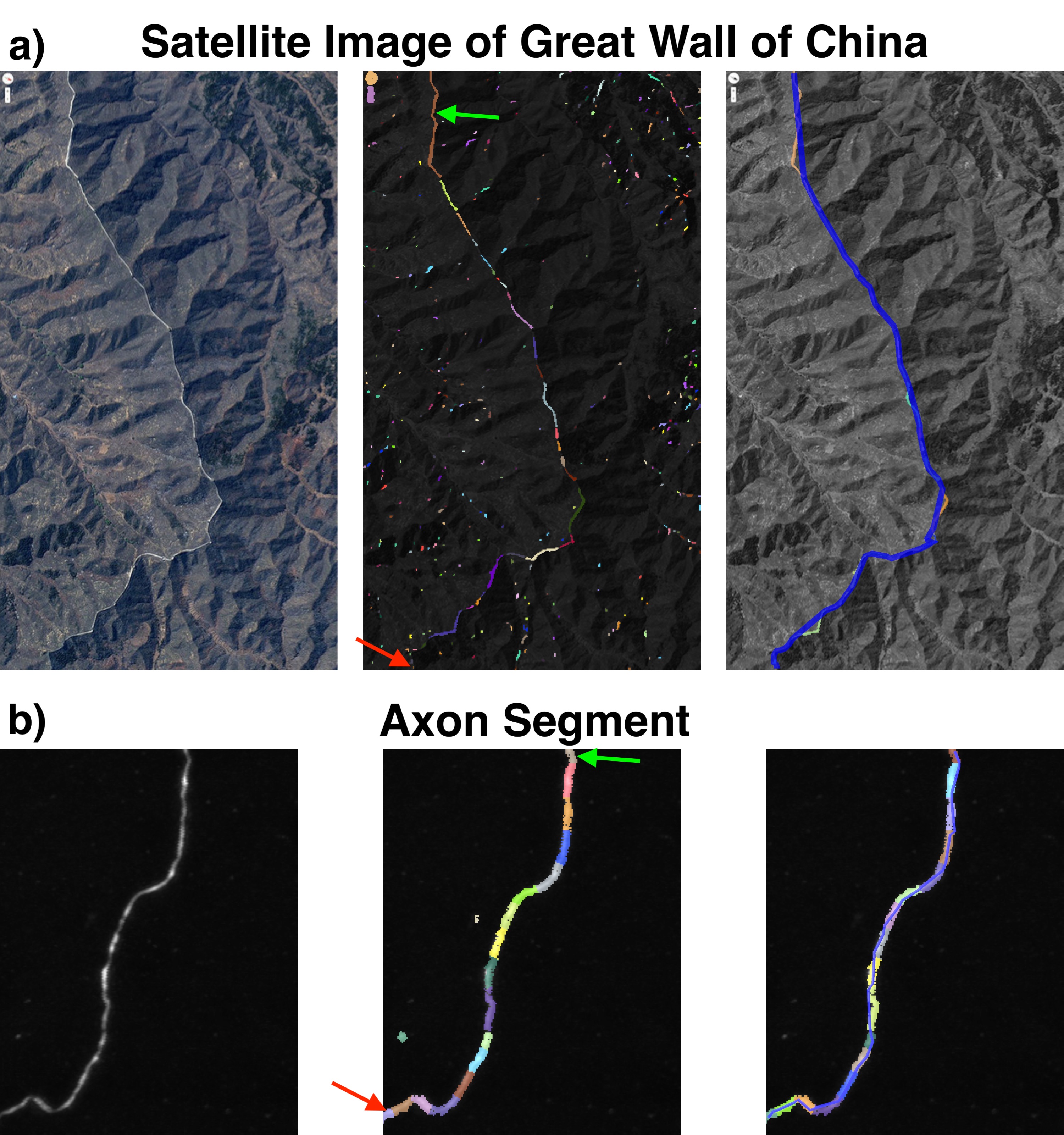}
    \caption{Demonstration of maximally probable reconstruction on isolated linear structures from a) a satellite image of part of the Great Wall of China and b) a neuronal process from the MouseLight dataset (MIP). Left panels show the original images. Middle panels shows the space of fragments, $\F$, pictured in color. The green and red arrows indicate the start and end states of the reconstruction task, respectively. The right panels show the most probable fragment sequences, where the fragments are colored and overlaid with a blue line connecting the endpoints of the fragments.}
    \label{fig:wall}
\end{figure}

Figure \ref{fig:wall} demonstrates the reconstruction method on both a satellite image of part of the Great Wall of China, and part of an axon. Different image segmentation models were used to generate the fragments in the two cases, but the process of joining fragments into a reconstruction was the same. The algorithm reconstructs the one dimensional structure in both cases. 

\begin{figure}[ht]
    \centering
    \includegraphics[width=1\textwidth]{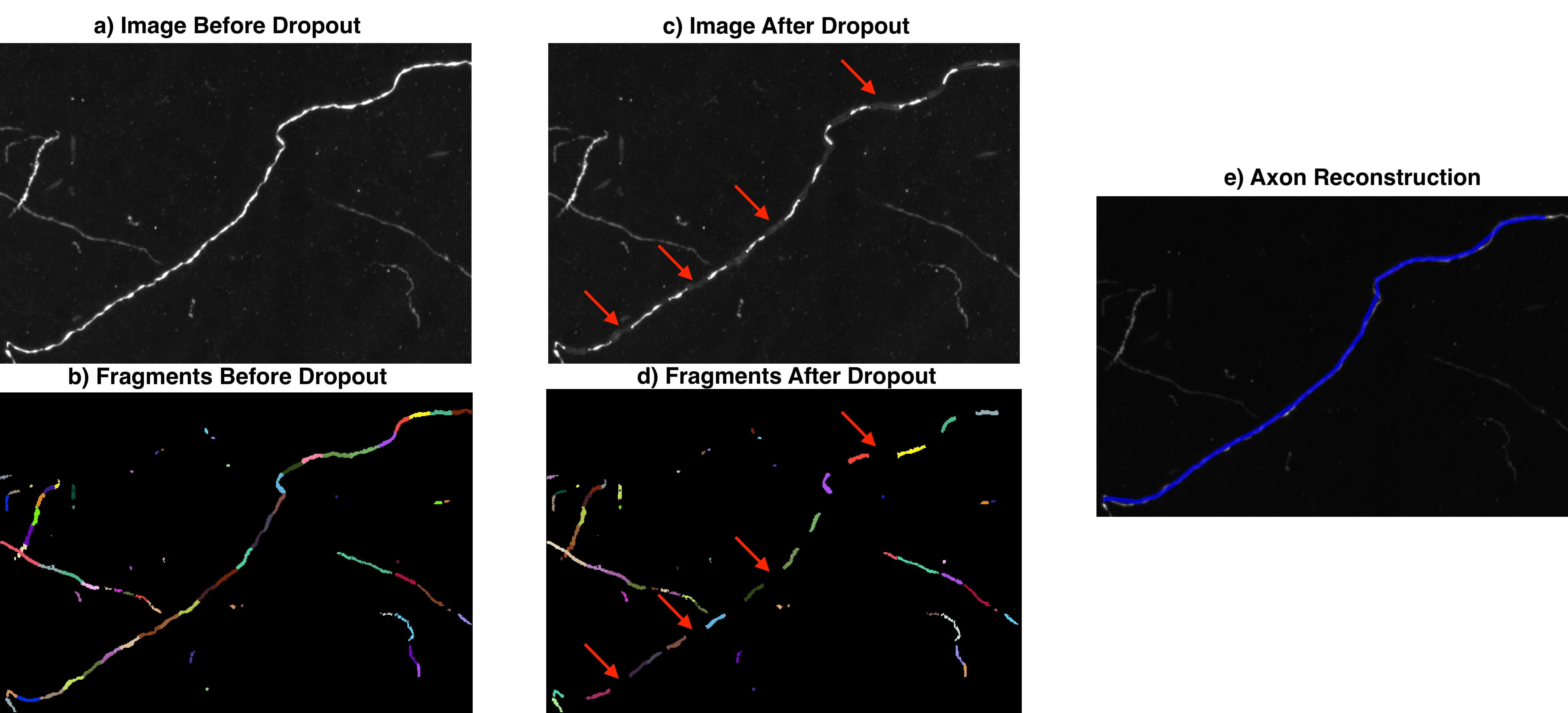}
    \caption{ViterBrain is robust to image intensity and fragment dropout when axons are relatively isolated. In this image, we censored the image intensity and the fragment space periodically along an axon path (red arrows in panels c and d show censored sections). Nonetheless, our algorithm was able to jump over the censored regions to reconstruct this axon (panel e). All images are MIPs}
    \label{fig:deletion-of-sections}
\end{figure}

Since the state transition probabilities are modeled with a %Boltzmann 
Gibbs distribution indexed by the state (hence giving the Hidden Markov property), reconstructions are driven by relative energies of different transitions. Thus, our method can still be successful in the presence of luminance or fragment dropout, as long as the neuronal process is relatively isolated (Figure \ref{fig:deletion-of-sections}). Our geometric prior has two hyperparameters, $\alpha_d$ and $\alpha_\kappa$, which determine the influence of distance and curvature, respectively, on the probability of connection between two neuronal fragments. 

\begin{figure}[ht]
    \centering
    \includegraphics[width=0.9\textwidth]{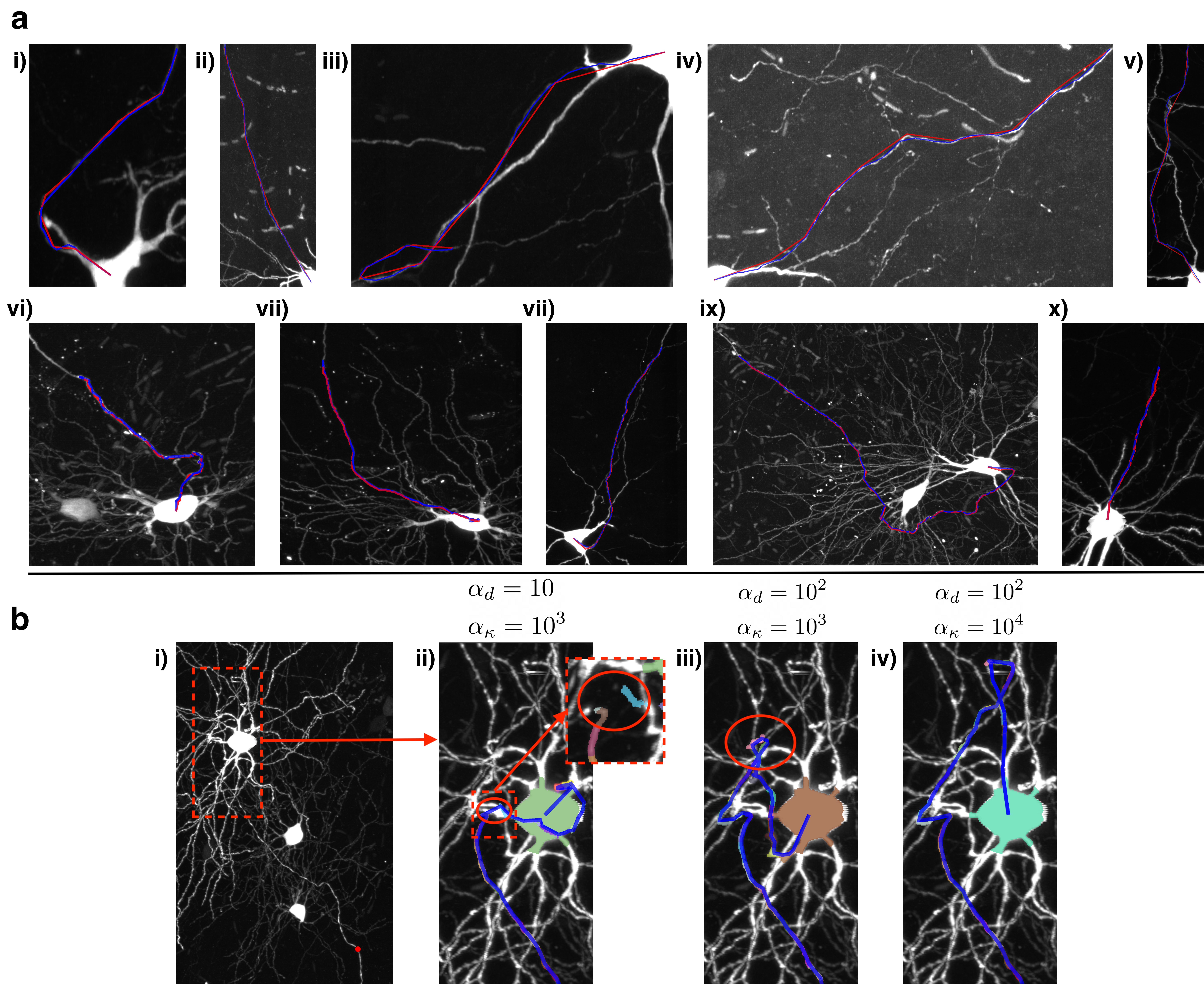}
    \caption{a) Successful axon reconstructions; the ViterBrain reconstructions are shown by the blue line; the manual reconstructions are shown by the red line. The algorithm was run with the same hyperparameters in each case: $\alpha_d = 10$ and $\alpha_\kappa = 1000$. 
    b)
    Different hyperparameter values lead to different results. Panel i) shows the neuron of interest. Panels ii-iv) are close-up views of reconstructions with different hyperparameter values that weigh transition distance ($\alpha_d$) and transition curvature ($\alpha_\kappa$). The red circle in Panel ii) indicates where the reconstruction deviated from the true path by jumping $\sim 10$ $\upmu$m to connect the gray fragment to the light blue fragment. Panel iii) shows how a higher $\alpha_d$ value avoids the jump in panel ii), but takes a sharp turn to deviate from the true path (red circle). Finally, in panel iv), the reconstruction avoids both the jump from panel ii) and the sharp turn from panel iii) and follows the true path of the axon back to the cell body. All images are MIPs.
    }
    \label{fig:success}
\end{figure}

Figure \ref{fig:success}a
shows various examples of maximally probable reconstructions.  The algorithm was run with the same hyperparameters in all cases: $\alpha_d = 10$ and $\alpha_\kappa = 1000$.

In some cases, reconstruction accuracy is sensitive to hyperparameter values (Figure \ref{fig:success}b). Higher values of $\alpha_d$ penalize transitions between fragment states with large gaps; higher values of $\alpha_\kappa$ penalize state transitions with sharp angles as measured by their discrete curvature. It is important to note that the transition distributions depend on the exact values of $\alpha_d$ and $\alpha_\kappa$, not just, for example, the ratio between them. We found $\alpha_d = 10$ and $\alpha_\kappa = 1000$ to be effective for the reconstructions shown throughout the paper, but these values should be adjusted according to the quality of fragments, and the geometry of the neurons being reconstructed.

\subsection{Signal Dropout Causes Censored Fragment States}

\begin{figure}[ht]
    \centering
    \includegraphics[width=\textwidth]{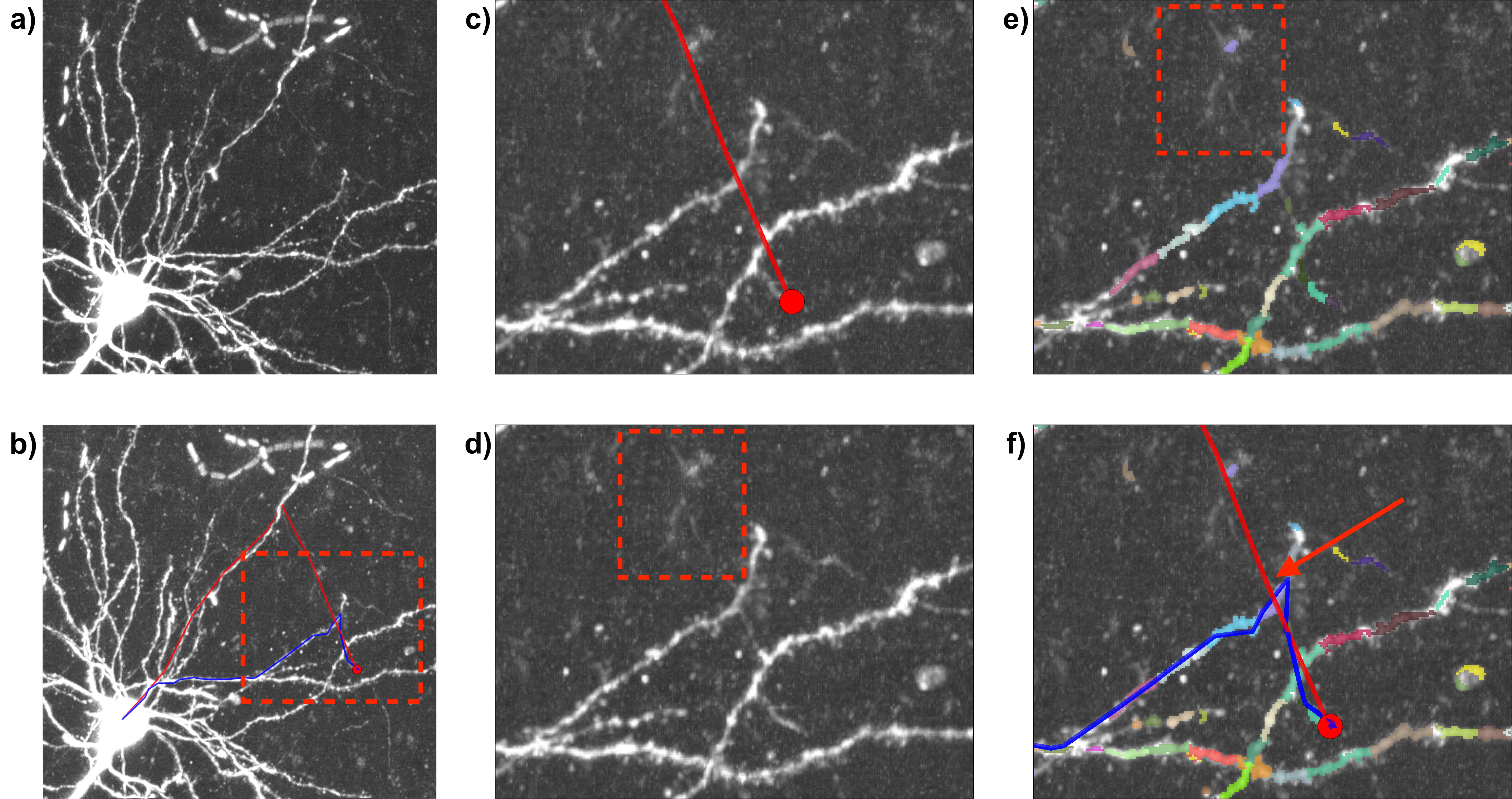}
    \caption{The most common failure mode is an inadequate fragment space due to signal dropout. 
    a) shows maximally probable reconstructions (blue lines) that deviate from the manual reconstructions (red lines).
    b) shows an example with an inadequate fragment space in more detail. The full image subvolume is shown in panel i) with the blue line depicting the most probable reconstruction and the red line depicting the manual reconstruction.
    The close up views in panels ii,iii) show that the image signal under the manual reconstruction is weak (red dotted box). Panel iv) shows the fragments in different colors. Indeed, there are only a few fragments generated in the region delineated by the red dotted box, where the manual reconstruction is. Panel v) shows the manual reconstruction in red and the algorithm's reconstruction in blue. The algorithm starts on the right path, then veers away from the true path at the red arrow. }
    \label{fig:bad_frags}
\end{figure}

Though our method is robust to fragment dropout in certain contexts, it is less robust when there is dropout near a parallel neuronal process (Figure \ref{fig:bad_frags}).
%In areas where there are many neuronal processes, fragment generation is key as it determines the adequacy of the states for generating maximally probably paths that are consistent.
Since the reconstruction is ultimately a sequence of fragments, missing fragments forces the algorithm to consider alternative paths thereby trading off the curvature of jumping to an adjacent state (nearby alternative) or continuing on course on what might appear to be the correct neuronal trajectory.

The most obvious source of fragment dropout is low image luminance,
%$I(y)$, 
leading to false negatives in the initial image segmentation. However, the reason for fragment dropout depends on the underlying segmentation model. Empirically, we found that the reconstruction algorithm fails when when the fragment generation process neglects portions that are greater than $\sim 10 \: \upmu$m in length.

\subsection{Comparison to State of the Art}
We examined the accuracy of ViterBrain compared to state of the art reconstruction algorithms. We identified four algorithms that have accompanying publications, and freely available open-source implementations. The first method is APP2 which starts with an oversegmentation of the neuron using a shortest path algorithm, then prunes spurious connections \citep{APP2}. The second is Snake which is based on active contour modeling \citep{Wang-Roysam-2011}. The third is called Advantra, based on the particle filtering approach by \citet{advantra-2019}. The final reconstruction software we use is GTree \citep{zhou2021gtree}, which uses the algorithm outlined in \citet{quan2016neurogps}. This algorithm is similar to APP2 in that it starts with an initial reconstruction that spans several neurons, then identifies false connections. APP2, Snake, and Advantra were both used with their default settings and hyperparameters in Vaa3D 3.2 for Mac. GTree version 1.0.4 was used on Linux. For GTree, the binarization threshold was set to $1.0$, which was qualitatively identified as a good threshold to capture the neuron. The default soma radius, $3\; \upmu$m, was used in the soma identification step.

\begin{figure}[ht]
    \centering
    \includegraphics[width=0.7\textwidth]{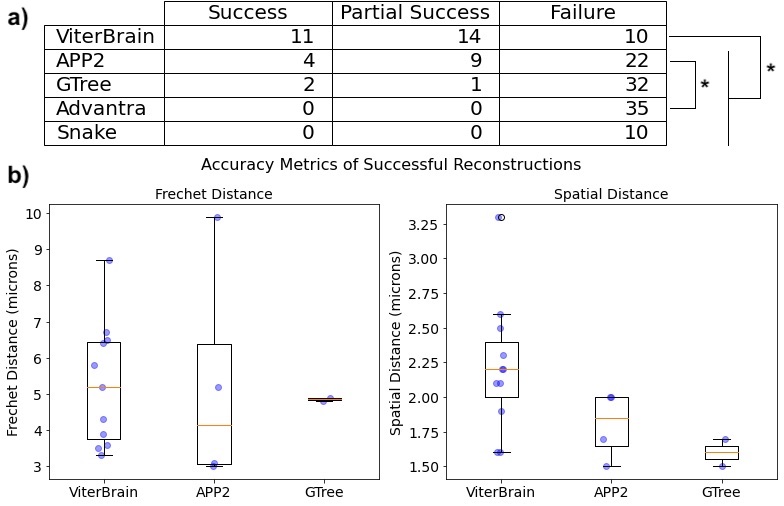}
    \caption{We applied several reconstruction algorithms to a dataset of 35 subvolumes of a MouseLight whole brain image (Snake was only applied to 10 subvolumes due to incoherent results and excessively slow runtimes, see Figure \ref{fig:snake}). Each subvolume contained a soma and part of its axon. The task was to reconstruct the portion of the axon that was contained in the image (no branching). First, the algorithms were evaluated visually and classified as successful, partially successful (over half, but not all, of the axon reconstructed), or failed. The table in panel a) shows these results, along with markers showing statistical significance in a two proportion z-test comparing success rates of the algorithms. For each successful reconstruction, we measured the Frechet distance and spatial distance from the manual ground truth in order to evaluate the precision of the reconstructions. These distances are shown as blue points in b), overlaid with standard box and whisker plots (center line, median; box limits, upper and lower quartiles; whiskers, 1.5x interquartile range; points, outliers).}
    \label{fig:results}
\end{figure}

\begin{figure}[ht]
    \centering
    \includegraphics[width=\textwidth]{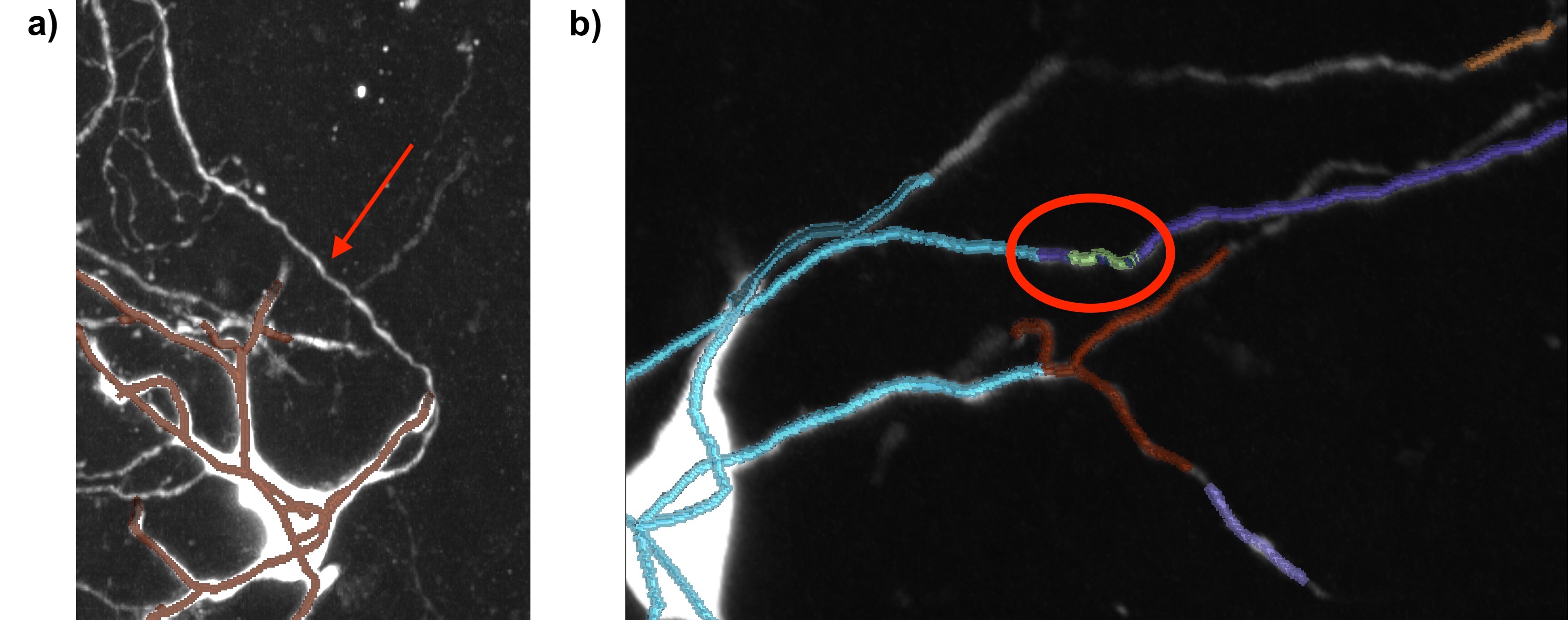}
    \caption{GTree produced sensible reconstructions on our dataset, but often fell short of fully reconstructing the axon segment in question. We observed two common failure modes with this algorithm. First, the reconstruction would sometimes fail to extend far enough down the axon. An example of this is shown in a) where the neuron reconstruction (brown overlay) does not extend up the axon identified with the red arrow. The other failure mode is when an axon would be split into several components such as in b). Each colored overlay in b), including green, blue, and purple represents a distinct reconstruction object produced by GTree. In other words, GTree severed the neuron (emphasized by the red circle) into several components and failed to fully reconstruct the axon.}
    \label{fig:gtree}
\end{figure}

\begin{figure}[ht]
    \centering
    \includegraphics[width=\textwidth]{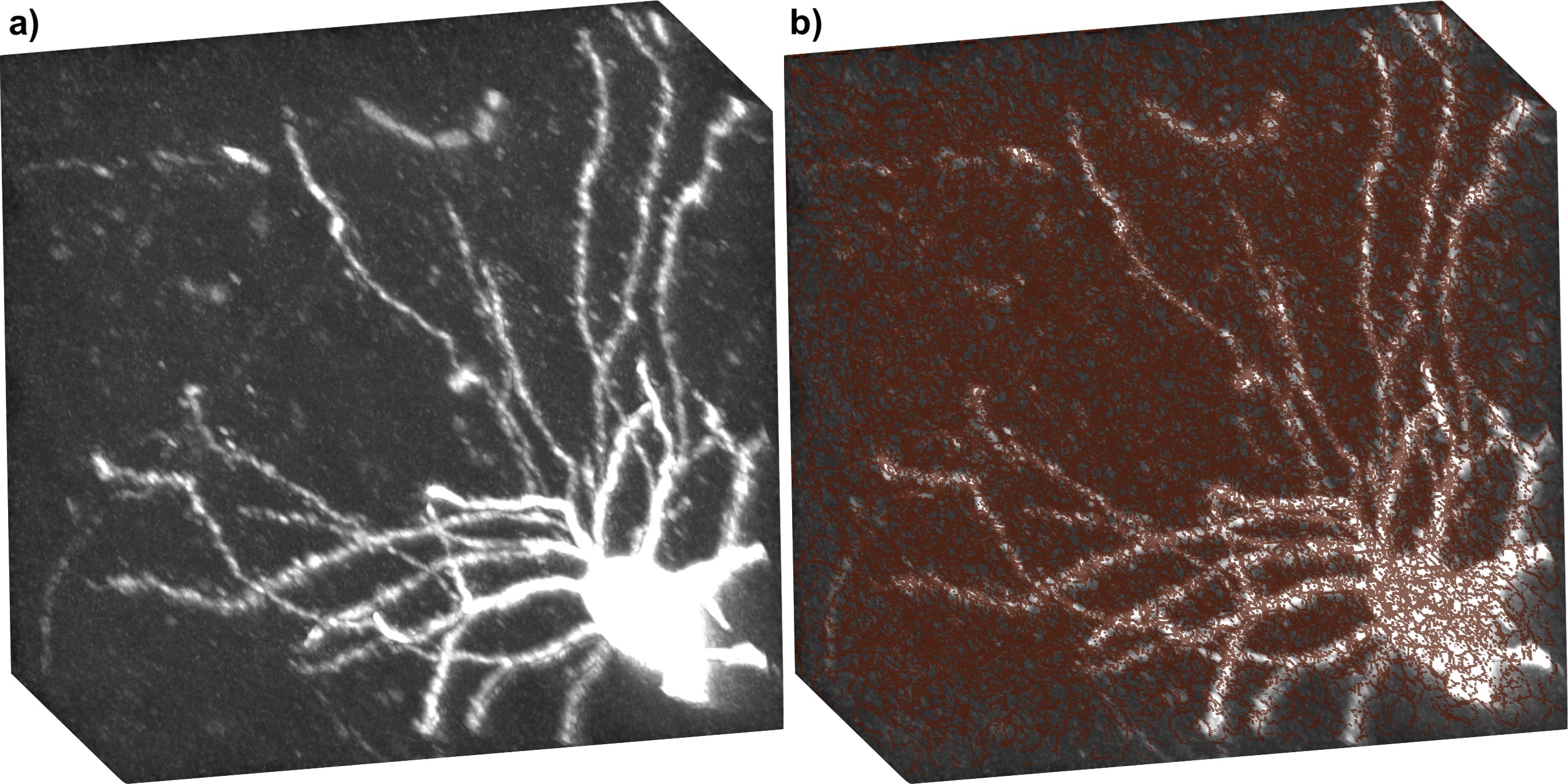}
    \caption{The Snake tracing algorithm produced incoherent reconstructions on our dataset. a) shows one of the image subvolumes, and the brown overlay in b) shows a reconstruction by Snake which is a dense, convoluted set of tangled paths. This algorithm was applied to 10 different subvolumes and the results were similar each time. There were no options to change any parameters in the Vaa3D implementation of this algorithm, and the algorithm took over an hour to process any subvolume larger than 50 megabytes, so we terminated its use after only 10 subvolumes.}
    \label{fig:snake}
\end{figure}

\begin{figure}[ht]
    \centering
    \includegraphics[width=\textwidth]{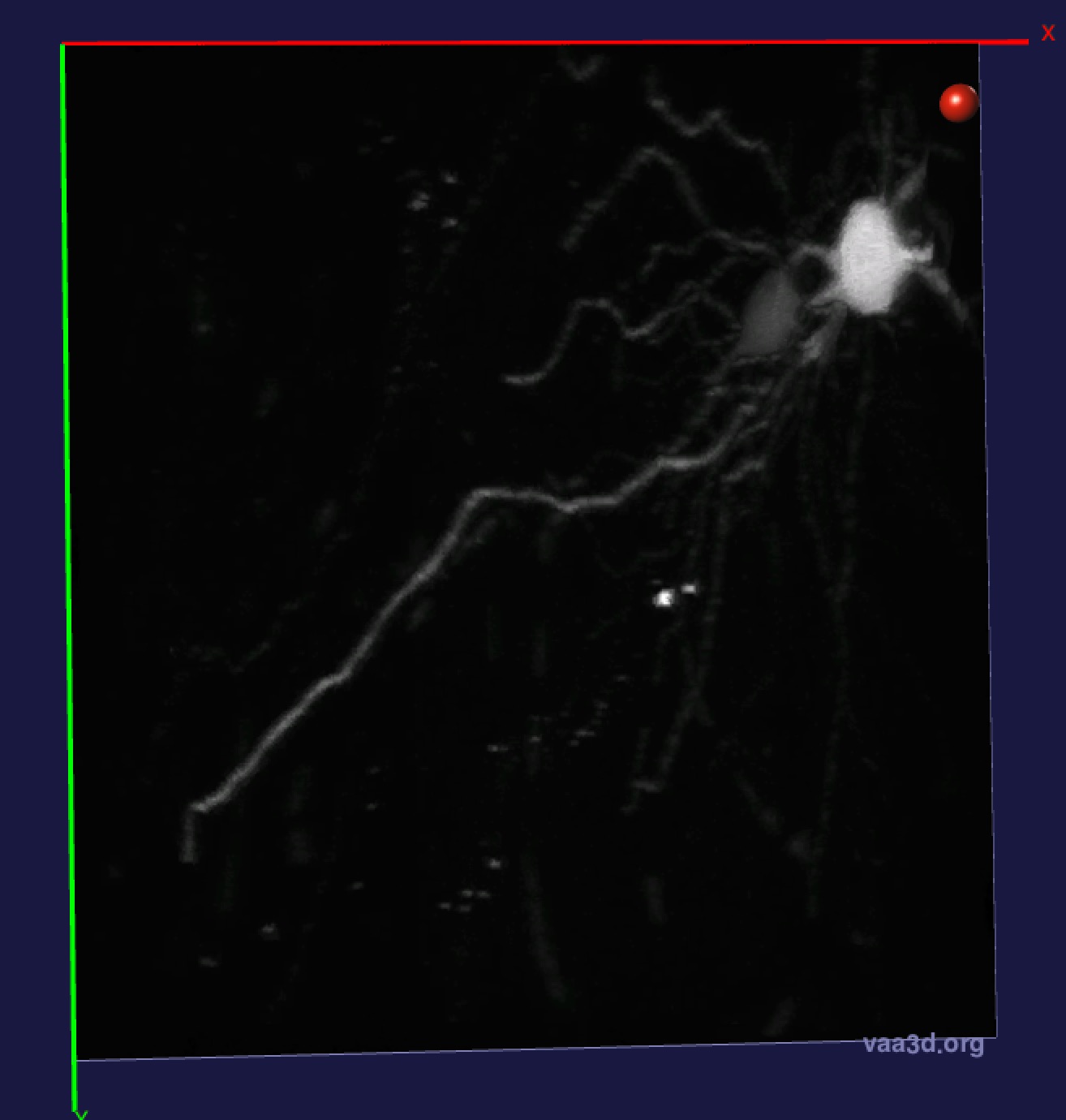}
    \caption{The Advantra algorithm, as implemented in Vaa3D, did not produce coherent reconstructions on the test dataset. Even under different hyperparameter settings, the algorithm would often produce a reconstruction composed of only a single point, like the one shown here.}
    \label{fig:advantra}
\end{figure}

Shown in Figure \ref{fig:results}a are the results of the various methods on a dataset of 35 subvolumes of a MouseLight whole brain image. Each subvolume contains a cell body, and the initial part of its axon that is covered by the first ten points of the Janelia reconstruction. So, the subvolumes vary in size but usually encompass around $10^6$ cubic microns. The algorithms are evaluated on how well they can trace the axon between fixed endpoints (cell body and tenth axon reconstruction point). The outcomes were classified as either successful (if the axon was fully traced), partially successful (if more than half of the axon was reconstructed as evaluated visually), or failures. According to two proportion z-tests the success rate of ViterBrain ($11/35$) was higher than all other methods at $\alpha=0.05$. Also, APP2 had a higher success rate ($4/35$) than Advantra at $\alpha=0.05$. The success rates for several of the algorithms are discouragingly low, so we discuss the possible reasons for this in the Discussion, and demonstrate the typical failure modes in Figures \ref{fig:gtree}, \ref{fig:snake} and \ref{fig:advantra}.

For the successful reconstructions, we examined the precision of the reconstructions using spatial distance (SD) and Frechet distance. We manually upsampled the Janelia reconstructions to provide the most precise ground truth. All reconstructions were sampled at every $1 \; \upmu$m before distances were measured. Spatial distance represents the average distance from a point on one reconstruction too the closest point on the other reconstruction \citep{peng2010v3d}. Frechet distance represents the maximum distance between two reconstructions, and satisfies the criteria to be a mathematical metric and is invariant to reparameterization (see section \ref{accuracy-metrics} for a derivation of Frechet distance). The spatial distance between ViterBrain and manual reconstructions are typically between 1 and 3 microns which is roughly the resolution of the image. Successful reconstructions by the other algorithms are also in this range. Frechet distances are larger, since they indicate maximal distance, not average distance, between curves. We observe that the $\sim 5$ micron deviations occur most often near the axon hillock where the axon broadens to merge with the soma.

\subsection{Proof of Concept Graphical User Interface}

\begin{figure}[ht]
    \centering
    \includegraphics[width=0.7\textwidth]{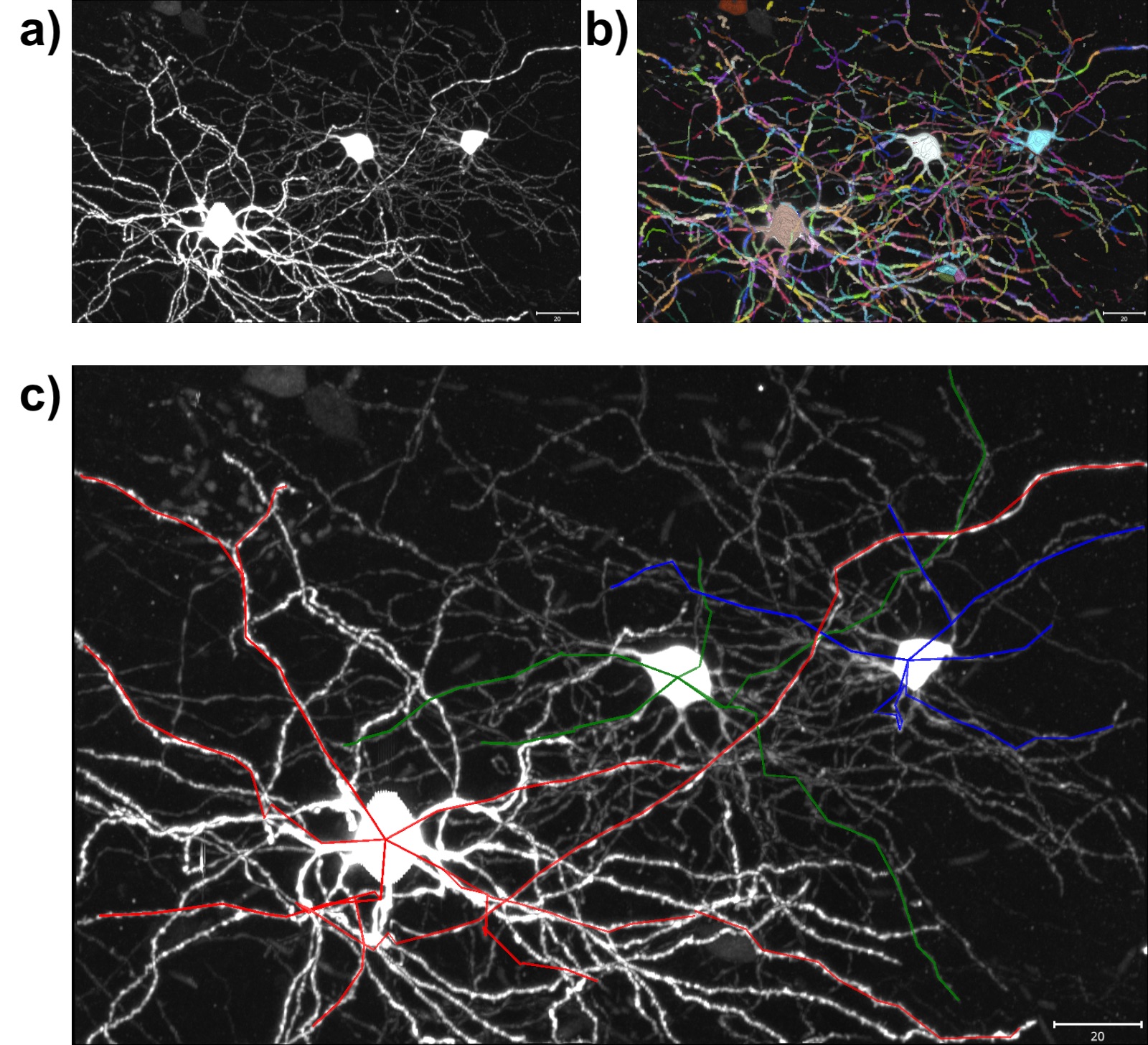}
    \caption{We designed a proof of concept graphical user interface where the user is presented with the image (a)) and the fragments (b)). The user can then click on two fragments and generate the most probable curve between them. c) shows three partial reconstructions (red, green and blue) of different neurons using the GUI.}
    \label{fig:octopus}
\end{figure}

Since our method is formulated as an optimal control problem conditioned on the start and states it is most suited for semi-automatic neuron reconstruction where a tracer could click on two different fragments along a neuronal process, and the algorithm would fill in the gap. The user could proceed to trace several neuronal processes until the full neuron is reconstructed. We designed a proof of concept graphical user interface based on this workflow and show an example set of reconstructions made using our GUI in Figure \ref{fig:octopus}. The GUI relies on the visualization software napari \citep{napari}.

\section{Discussion}
This paper presents a hidden Markov model based reconstruction algorithm that connects fragments generated by appearance modeling. Our method converts an image mask into a set of fragments and thus can be applied to the output of an image segmentation model. We chose Ilastik to generate image masks because of its convenient graphical user interface, and high performance on a small number of samples \citep{berg2019ilastik}. However, masks could also be generated using a deep learning based model such as \citet{liu2018improved, li20193d} or \citet{wang2021voxel}. 

These fragments are assembled based on the associated appearance model score of the observed image and the discrete numerical curvature and distance of adjacent fragments. In the Methods, we derive the Bayes posterior distribution of the hidden state sequence encoding the axon reconstruction. We also show that applying the polynomial time Viterbi algorithm is not possible for maximum a-posteriori estimation since the state space has to grow to account for possible cycles in the unordered image domain. We address the problem with cycles with a modified procedure for defining the path probabilities making it feasible to efficiently calculate the globally optimal neuron path.
The solution for efficiently generating the globally optimal path implies that the local minimum associated to gradient and active appearance models solutions is resolved in this setting.

We apply the algorithm to the fixed endpoint problem in two-photon images of mouse neurons. In a dataset of 35 partial axons, our algorithm successfully reconstructs more axons than existing state of the art algorithms (Figure \ref{fig:results}). We observed that the most common failure mode in this dataset was when there are extended ($>\sim 10$ micron) stretches of the putative axonal path where
there is significant loss of luminance signals leading to highly censored fragment generation. This implies that the maximally probable HMM procedure is only effective if paired with effective voxel classification tools. The algorithm can also fail in areas densely populated with neuronal processes. We demonstrate that proper selection of the hyperparameters to reflect the density of the fragments and the geometry of the underlying neurons can resolve these issues. Our algorithm is specifically adapted for reconstructing axons in projection neurons in datasets such as MouseLight in two ways. First, the high image quality as indicated by large KL-divergence values (Figure \ref{fig:fgbg}a), makes it straightforward to build an effective foreground-background classifier which is an essential part of our HMM state generation. Secondly, our algorithm encodes the geometric properties of axons such as curvature, which allows our solutions to adapt to the occasional sharp turns in projection axons.

The success rates of the other algorithms on our dataset is quite low, considering the performances that they achieved in their accompanying publications. There are two likely reasons for this, dataset differences, and sub-optimal algorithm settings.

When an algorithm is validated on one type of data, the results do not necessarily hold for datasets of a different type. Since none of the existing algorithms had been designed for the MouseLight data, the unique details of our dataset could lead to reduced performance. For example, several subvolumes in our dataset contain multiple neurons, while two of the algorithms, APP2 and Advantra, were explicitly designed for images containing single neurons. Snake is designed to handle multiple neurons, but is largely validated on DIADEM, a single neuron dataset \citep{peng2015diadem}. Other dataset differences include different image resolutions (or levels of anisoptropy), different image encodings (8 bit vs. 16 bit), and different signal to noise ratios. Lastly, the existing algorithms are designed to reconstruct all dendrites and axons simultaneously while our task of reconstructing a single section of the axon does not give the algorithms credit for successfully reconstructing other parts of the neuron. 

It is also important to note that reconstruction algorithms can be sensitive to hyperparameter settings. All algorithms had different hyperparameter options except for Snake. While we tried various hyperparameter settings for all algorithms, the only non-default setting that clearly improved reconstruction performance was the binarization threshold setting in GTree, all other settings we left to default. It is likely that these settings were not optimal for our dataset, but it is quite time intensive for a typical user to quantitatively determine the optimal settings. Detailed and accessible software documentation makes the process of choosing effective algorithm settings more efficient.

Figures showing the common failure modes for some of the algorithms are shown in the Supplement \ref{sec:supp-figs}. Advantra and Snake produced incoherent reconstructions on the dataset, and the common failure modes for GTree were early termination of reconstruction, or severing an axon into multiple components. Despite the possible reasons for sub-optimal performance of the other algorithms, we provide evidence that our algorithm is competitive with state of the art methods for reconstructing neuronal processes. Future benchmark comparisons could include reinforcement learning, or recurrent neural network approaches, which have become prevalent in sequential decision processes. However, there is not much scientific literature on these approaches to neuron reconstruction with accompanying functional code.

\begin{figure}[ht]
    \centering
    \includegraphics[width=\textwidth]{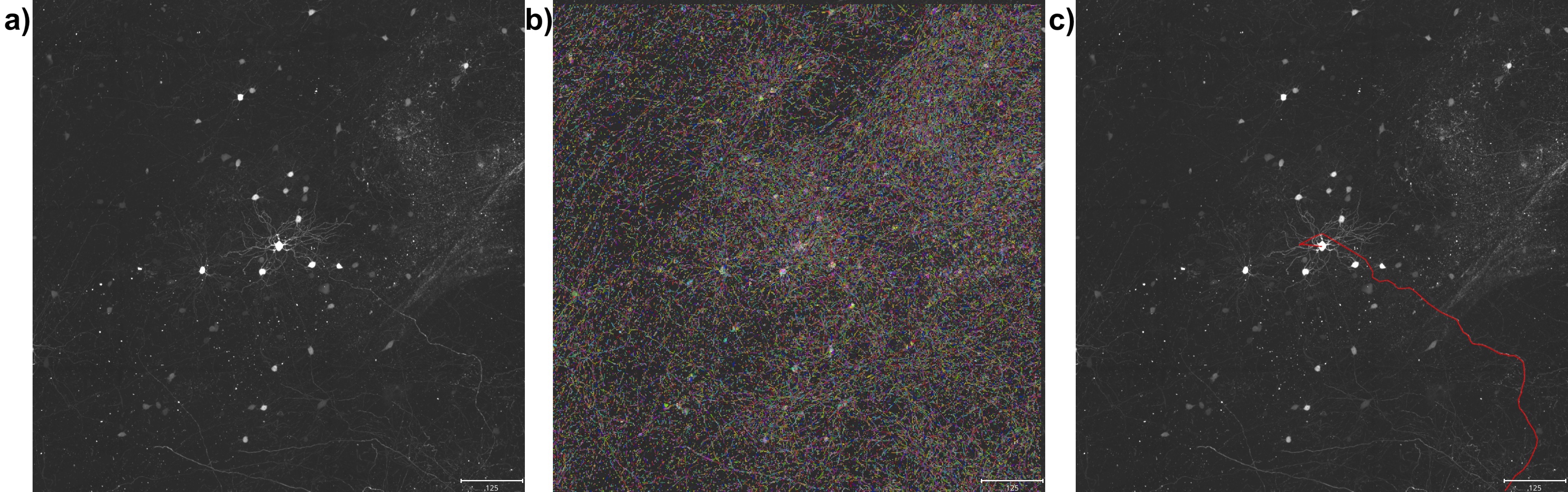}
    \caption{We applied the ViterBrain pipeline to an image subvolume with dimensions $3332\times 3332 \times 1000$ voxels, which encompasses a cubic millimeter of tissue. Pictured is a set of maximum intensity projections of a downsampled version of this subvolume. a) shows the image, b) shows the image with a color overlay depicting the neuron fragments and c) shows a trace in red that follows an axon. The scale bar represents $125$ microns.}
    \label{fig:1mm}
\end{figure}

We have scaled up the ViterBrain pipeline to process images of $3332\times 3332\times 1000$ voxels, representing one cubic millimeter of tissue (Figure \ref{fig:1mm}), and we will continue to improve the pipeline until it can be run on whole brain images. The traces generated by our pipeline could also be paired with tools like the one presented in \citet{li2020brain} which can turn traces into full neuron segmentations complete with axon and dendrite thickness measurements.

The code used in this work is available in our open-source Python package \texttt{brainlit}: \\ http://brainlit.neurodata.io/, and a tutorial on how to use the code is located at: http://brainlit.neurodata.io/link\_stubs/hmm\_reconstruction\_readme\_link.html.

\section{Methods}
\label{sec:methods}
\subsection{The Bayesian Appearance Imaging Model}
\label{sec:appearance}
Our Bayes model is comprised of a prior which models the axons as geometric objects and a likelihood which models the image formation process.

We model the axons as simply connected curves in $\R^3$ written as a function of arclength
$$ c(\ell), \ell \geq0 \ , c(\ell) \in \R^3 \ .$$
We denote the entire axon curve in space as $c(\cdot) \coloneqq \{c(\ell), \ell \geq 0\}$.
To interface the geometric object with the imaging volume
we represent the underlying curve $c(\cdot)$ as a delta-dirac impulse train in space. %viewed as a generalized function $C(y) = \int_{0}^1 \delta (y-c(\ell)) d \ell $.
We view the imaging process as the convolution of the delta-dirac impulse train with the point-spread kernel of the imaging platform.
%$$ \kappa(y;c(\cdot)) =k(y)  \Asterisk \delta(y-c(\cdot)) \ . $$
We take the point-spread function of the system to be roughly one micron in diameter, implying that the axons are well resolved. 
The fluorescence process given the axon contour %$\kappa(\cdot; c) $ 
is taken as a relatively narrow path through the imaging domain ($\sim  1\;\upmu$m diameter) with relatively uniform luminance.

%\subsection{The Appearance Foreground-Backround Model.}
We take the image to be defined over the voxel lattice $ D=\cup_{i \in \Z^{m^3}} \Delta y_i \subset \R^3 $ with centers $y_i \in \Delta y_i$. We model the image as a random field 
$\{I_{y_i}, \, \Delta y_i \in D\}$ whose elements are independent when conditioned on the underlying axon geometry, similar to an inhomogeneous Poisson process as described in \cite{snyder2012random}. We denote the image random field associated to any subset of sites $Y \subset D$,  with joint probability conditioned on the axon:
\begin{subequations}
\begin{align}
&I_{Y}= \{ I_{y_i}: \Delta y_i \in Y\} \ ;
\\
&P(I_Y| c(\ell), \ell \geq 0) = \prod_{y \in Y} p(I_y| c(\ell), \ell \geq 0) \ .
\end{align}
\end{subequations}
Because of the conditional independence, the marginal distribution determines the global joint probabilities. Of course, while the conditional probabilities factor and are conditionally independent, the axon geometry is unknown and the measured image random field is completely connected if the latent axon process is removed.
We adopt a two hypothesis formulation $\{f,b\}$ corresponding to a foreground-backround model for the images where the marginal probability of a voxel intensity is:
\begin{subequations}
\begin{align}
\label{foreground-backround-equation}
&
P(I_{y}) = \alpha_1(I_{y}) , y \in \text{foreground} \\
&P(I_{y}) = \alpha_0(I_{y}) , y \in \text{background}
\label{backround-equation}
\end{align}
\end{subequations}

Our conditional independence assumption implies that the joint distribution of group of foreground or group of backround voxels can be decomposed into the corresponding marginal distributions over the foreground-backround models \eqref{foreground-backround-equation},\eqref{backround-equation}; defining the foreground and backround sets $Y = Y_f \cup Y_b$, then 
\begin{subequations}
\begin{align}
%& \ell(I_{F}; F) = \prod_{y \in F} \alpha_1 (I_{y}) \ .
%& P(I_Y ) = \prod\nolimits_{y \in Y} \alpha_1 (I_{y}), \  Y \subset \text{foreground} \ ,\\
& P(I_Y ) = \prod\nolimits_{y \in Y_f } \alpha_1 (I_{y})\, \prod\nolimits_{y \in Y_b} \alpha_0 (I_{y})
\ . % \ Y= Y_0 \cup Y_1= %\text{foreground} \ ,
%\\ & P(I_Y ) = \prod\nolimits_{y \in Y} \alpha_0 (I_{y}), \  Y \subset \text{background} \ ;
\label{Poisson-likelihood-representation}
\intertext{We define the notations for the joint probability
of the set in the foreground for example (or background)}
&
%\ell(I_{F_1},\dots, I_{F_n};F_1,\dots, F_n) =
\alpha_1(I_{Y_f}) \coloneqq
\prod\nolimits_{y \in Y_f} \alpha_1(I_{y}), 
 \ Y_f \subset  \text{foreground} \ .
 \label{eq:alpha-product-probability}
\end{align}
\end{subequations}

Despite the Poisson nature of the image acquisition process, simple scaling or shifting of the imaging data would mean the image intensities are no longer Poisson. To accommodate this effect, we estimate the foreground-background intensity distributions ($\alpha_1(\cdot)$ and $\alpha_0(\cdot)$ respectively) nonparametrically. The simplest nonparametric density estimation technique is using histograms. However, it can be difficult to choose the origin and bin width of histograms, so we opt for a kernel density estimate (KDE) approach. We estimate $\alpha_0(\cdot)$, $\alpha_1(\cdot)$ by labeling a subset of the data as foreground/background then fitting Gaussian KDEs to the labeled data (see Figure \ref{fig:fgbg}b). We use the scipy implementation of Gaussian KDEs \citep{2020SciPy-NMeth}, with Scott's rule to determine the bandwidth parameter \citep{scott2015multivariate}. Under some assumptions on the derivatives of the underlying density, our approach converges to the true density as the number of samples increases (Theorem 6.1 in \cite{scott2015multivariate}). Further, the Scott's rule choice for bandwidth is (approximately) optimal with respect to mean integrated square error.

Our approach allows us to empirically estimate the intensity distributions while maintaining the independent increments property of a spatial Poisson process, which is consistent with
the autocorrelation curves depicted in Figure 
\ref{fig:fgbg}a. This is the key property that allows us to factor joint probabilities into products of marginals for sets of voxels.

We note that the foreground-background imaging model allows us to estimate the error rate of classifying a voxel as either foreground or background. In the Neyman-Pearson framework, foreground-background classification is a simple two hypothesis testing problem and the most powerful test at a given type 1 error rate is the log likelihood ratio test. The Kullback-Leibler (KL) divergence between the foreground and background distributions gives the exponential rate at which error rates converge to zero as the number of independent, identically distributed samples increases \citep{cover1991elements}. In the case of using Gaussians to model the foreground-background distributions, the KL divergence reduces to the squared signal to noise ratio. In the absence of normality, the KL Divergence is the general information theoretic measure of image quality for arbitrary distributions. We propose KL Divergence as an important statistic in evaluating the quality of fluorescent neuron images.

\subsection{The Prior Distribution via Markov State Representation on the Axons Fragments}
\label{sec:prior}
\label{sec:frags}

Our representation of the observed image is as a hidden Markov random field with the axon as the hidden latent structure.
Given the complexity of sub-micron resolution images, we build an intermediate data structure at the micron scale that we call \textit{fragments} $F  \subset D$ 
defined as collections of voxels without any assumed global ordering between them. Each fragment represents a portion of a neuronal process, with a natural orientation given by one end that is closer to the soma.
Depicted in Figure \ref{fig:components} are the fragments shown via different colors. 
The fragments are a coarser scale voxel and can be viewed analogously as higher order features.

The axon reconstruction problem becomes the reassembly of the fragments along with the imputation of the censored fragments.
In the image examples presented in this work, the complexity of the space of fragments is approximately $|\F| = 100$,$000$ for a cubic millimeter of projection neuron image data.
%The probability of a fragment under the foreground-background appearance model is the simple product probability
%from Eq. \eqref{eq:alpha-product-probability},  for $F \subset \text{foreground}$, 
%$ \alpha_1(I_{F})=\prod\nolimits_{y \in F} \alpha_1(I_y)$, $k=0,1 $.  

We exploit the computational structures of hidden Markov models (HMMs) when the underlying latent structure is absolutely ordered so that dynamic programming can efficiently compute globally optimal state sequence estimates. From the set of fragments we compute a set of \textit{states} for the HMM. The states are a simplified, abstract representation of the fragments that contain the minimum information required to specify the HMM. Each state includes endpoints $x^0, x^1 \in \R^3$ in order to compute "gap" or "censored" probabilities, and unit length tangents $\tau^0, \tau^1 \in \R^3$ associated with the endpoionts in order to compute curvature. Each fragment generates two states, one for each orientation. The two states are identical except their endpoints are swapped, and their tangents are swapped and reversed. We denote the natural mapping from state to fragment by $F:s \in \S \mapsto F(s) \in \F$.

The collection of states  $\S=\{s\}$ is the finite state space of the HMM, and our goal is to estimate the state sequence $(s_1,...,s_n)$ that follows the neuronal process.

Our algorithms exploit two splitting properties, the Markov nature of the state sequence and the splitting of the random field image conditioned on the state sequence. We use the notation $s_{i:j}:=(s_i,s_{i+1},\dots, s_{j}) $ for partial state sequences.
We model the state sequence $s_1,\dots, s_n$  as Markov with splitting property:
$$
p(s_{i+1:n}, s_{1:i-1} | s_i) = p( s_{i+1:n}| s_i) p(s_{1:i-1}|  s_i) \ , \ i =2, \dots, n-1 \ ,
$$
which implies the 1-order Markov property $p(s_i|s_{i-1}, s_{1:i-2}) = p(s_i|s_{i-1}) $.
%with $p(s_1|s_0) = \pi(s_1)$ the initial distribution.

We define the transition probabilities with a Boltzmann distribution with energy $U$:
\begin{align*}
    p(s_i | s_{i-1}) & = \frac{e^{-U(s_{i-1}, s_i)}}{Z(s_{i-1})} \ , \ \text{with} \ Z(s_{i-1}) = \sum_{s_i \in \S} e^{-U(s_{i-1}, s_i)}  \ ,
\\
\intertext{with energy given by}
    U(s_{i-1}, s_i) &= \alpha_d |x_{i}^0 - x_{i-1}^1|^2 + \alpha_\kappa \kappa(s_{i-1}, s_i)^2 \ ,
\end{align*}
where $|\cdot|$ is the standard Euclidean norm. The two hyperparameters $\alpha_d$ and $\alpha_\kappa$, determine the influence of distance and curvature, respectively, on the probability of connection between two neuronal fragments. The term $\kappa(s_{i-1}, s_i)$ approximates the curvature of the path connecting $s_{i-1}$ to $s_i$ as follows:

Define
$\tau_c \coloneqq \frac{ x_{i}^0 - x_{i-1}^1}{||x_{i}^0 - x_{i-1}^1||}$ which is the normalized vector connecting $ s_{i-1}$ to $ s_i $. Then we can approximate the squared curvature at $x_{i-1}^1$ and $x_{i}^0$ with $\kappa_1(s_{i-1}, s_i)^2 = 1 - \tau_{i-1}^1 \cdot \tau_c$ and $\kappa_2(s_{i-1}, s_i)^2 = 1 - \tau_c \cdot \left( -\tau_i^0\right)$ respectively. These formulas are derived in Supplement \ref{appendix-curvature-approximation} and they approximate curvature as modeled in \citet{athey2021spline}. Finally, we approximate average squared curvature with the arithmetic mean:

\begin{align*}
    \kappa(s_{i-1}, s_i)^2 = \frac{\kappa_1(s_{i-1}, s_i)^2+\kappa_2(s_{i-1}, s_i)^2}{2}
\end{align*}

We control the computational complexity
associated to computing prior probabilities for all $|\S|^2$ states by restricting the possible state transitions. We set to 0 probability any transitions where the distance between endpoints is greater than $15 \, \upmu$m or the angle between states is greater than $150^\circ$. We also set the probability that a state transitions to itself to zero.

\subsection{Global Maximally Probable Solutions via %Viterbi 
$O(n|\S|^2) $ Calculation}
%\subsection{The MAP State Sequence}
The maximum a-posteriori (MAP)
state sequence
is defined as the maximizer of the posterior probability (MAP) over the state sequences $s_{1:n} \in {\S}^n$, with $|\S|$ finite:
\begin{equation}
\hat s_{1:n} := \argmax_{s_{1:n} \in {\S}^n} \log p(s_{1:n} |  I_D) \ . %=\argmax_{s_{1:n} \in {\S}^n} \log p(s_{1:n} |   I_{D\setminus \cup_{i=1}^n F_i},\, F_{1:n}) \ .
\label{MAP-estimator-equation}
\end{equation}

The solution space has cardinality $|\S|^n$ so it is infeasible to compute the global maximizer by exhaustive search. Our approach is to rewrite the probability recursively in order to use the Viterbi algorithm and dynamic programming with $O(n|\S|^2)$ time complexity.
We rewrite the MAP estimator in terms of the joint probability:
$$\hat s_{1:n} = \argmax_{s_{1:n} \in \S^n}  p(s_{1:n} |  I_D) %= \argmax_{s_{1:n} \in \S^n}  p(s_{1:n} ,  I_D) 
= \argmax_{s_{1:n} \in \S^n}  p(s_{1:n} , I_{D})
 \ . 
$$
The image random field is split or conditionally independent conditioned on the fragment states:
%Given the state, the field is conditionally indpendent 
$$p(I_{F(s_i)}, I_{D\setminus F(s_i)}| s_i, )= p(I_{F(s_i)}| s_i) p(I_{D\setminus F(s_i)}| s_i) \ ,
$$
which implies $p(I_{F(s_i)}| s_i,  I_{D\setminus F(s_i)}) =p(I_{F(s_i)}| s_i) $.

%Let $y \in F \in \F$ be the set of voxels associated to fragment $F$, then the Poisson model gives us the independent increments across the fragment and the "foreground probability" $\alpha_1$.
%The likelihood of the images associated to the fragment representation of the contour factors as above:
%$$ \ell (I_D; F_1,\dots, F_n) = \prod_{i=1,\dots,n} \prod_{y \in F_i }\frac{\alpha_1 (I_{y})}{\alpha_0 (I_{y})} \prod_{y \in D} \alpha_0 (I_{y}) \ . $$

Define the indicator function $\delta_A(x) =1$ for $x \in A$, $0$ otherwise. % and convention $x_{1:0} = \emptyset$.
\begin{lemma}
\label{Lemma-posterior-probability}
Defining the shorthand notation
%$ F_i \coloneqq F(s_i) $ 
identifying fragment sequences with the state sequence: $$ %F(s_{1:n}) \ \
F_{1:n} \coloneqq (F(s_1),\dots, F(s_n)) \ , $$
then for $n > 1$ we have the joint probability:
\begin{align}
%&p(s_{1:n}, I_{D}) =\left(\frac{\alpha_1(I_{F_n})}{\alpha_0(I_{F_n})}\right)^{\delta_{D\setminus F_{1:n-1}}(F_n)} p(s_{n}|s_{n-1} ) \,p(F_{1:n-1}, I_{D}) ,\label{recursion}
  p(s_{1:n}, I_{D}) &= \prod\nolimits_{i=2}^n 
  \left(
  %\prod\nolimits_{y \in F_i} 
  \frac{
  \alpha_1(I_{F_i})}{\alpha_0(I_{F_i})}\right)^{\delta_{D\setminus F_{1:i-1}}
(F_i)} p(s_{i}|s_{i-1}) \ p(s_1, I_{D}) 
\ . \label{eq:joint} 
\end{align}
\end{lemma}
See Supplement  \ref{appendix-proof-posterior-probability} for the proof
which rewrites the probability recursively giving the factorization. The proof is similar to the classic HMM decomposition in how it uses the two splitting properties, but there are two differences. The first is that the probability needs to account for the full image, including areas outside of the axon estimate, which explains the presence of both $\alpha_1$ and $\alpha_0$ in Eq. \ref{eq:joint}. The second difference is that if the state sequence $s_{1:n}$ contains repeated states, then the corresponding image data should not be double counted in the probability. This is enforced by the delta function $\delta(\cdot)$.

It is natural to take the negative logarithm of Eq. \ref{eq:joint} to obtain a sum that represents and the cost of a path through a directed trellis graph \citep{forney1973viterbi}. Several algorithms exist that solve the shortest path problem in $O(n |\S|^2)$ complexity. However, we cannot use these algorithms directly because the cost function is not ``sequentially-additive'' due to the dependence of the indicator function on previous states in the sequence.
%The state dependence grows without bound and therefore so does the complexity because of the indicator function.
%The recursion \eqref{recursion} shows it immediately, taking the negative log- probability:
%\begin{equation}
%p(s_{1:n}, I_{D}) =\left(\frac{\alpha_1(I_{F_n})}{\alpha_0(I_{F_n})}\right)^{\delta_{D\setminus F_{1:n-1}}(F_n)} p(s_{n}|s_{n-1} ) \,p(s_{1:n-1}, I_{D}) ,\label{recursion} \end{equation}
%\begin{align*}   
%- \ln p(s_{1:n}, I_{D}) &= - \left( \delta_{D\setminus F_{1:n-1}} (F_n) \ln \frac{\alpha_1(I_{F_n})}{\alpha_0(I_{F_n})} + \ln p(s_{n}|s_{n-1}) \right) -\ln p(s_{1:n-1},I_D) \ .
%\end{align*}
In the Supplement, section \ref{viterbi-counter-example}, we offer a example demonstrating that directly applying the Viterbi algorithm to this problem does not generate the MAP estimate.

We adjust our probabilistic representation on the $|\S|^n$ paths in order to utilize shortest path algorithms such as Bellman-Ford or Dijkstra's \citep{bellman1958routing, dijkstra1959note}.
For this we note that the $\frac{\alpha_1(I_{F_i})}{\alpha_0(I_{F_i})}$ term in Eq. \eqref{eq:joint} may often be greater than 1. In the directed graph formulation (negative logarithm transformation), this can lead to negative cycles in the graph of states. When negative cycles exist, the shortest path problem is ill-posed. To avoid this phenomena we remove the background component of the image from the joint probability, which converts $\frac{\alpha_1(I_{F_i})}{\alpha_0(I_{F_i})}$ to $\alpha_1(I_{F_i})$,
and converts our global posterior probability to our path probability formulation.

\begin{statement}
\label{maximum-likely-state-sequences}
Define the most probable solution $s_{1:n} \in \S^n$ by the joint probability
%\begin{equation}
$\argmax_{s_{1:n} \in \S^n} p(s_{1:n}, I_{F_{1:n}})$.
%\end{equation}
Then we have
\begin{equation}
\max_{s_{1:n} \in \S^n} p(s_{1:n}, I_{F_{1:n}} ) = \max_{s_{1:n} \in \S^n} \prod\nolimits_{i=2}^n  \left( 
%\prod\nolimits_{y \in F_i} 
\alpha_1(I_{F_i})
\right)^{\delta_{D \setminus F_{1:i-1}}(F_i)} p(s_i|s_{i-1}) \,
p(s_1, I_{F_1}) 
\ .
\label{eq:numerator_only}
\end{equation}

Further, if $\alpha_1(I_{y})\leq 1$ for all $y$, then the globally optimal solution to the fixed start and end point problem is a nonrepeating state sequence and can be obtained by computing the shortest path in a directed graph where the vertices are the states, and  the edge weight from state $s_{i-1}$ to $s_i$ is given by:

\begin{align}
    e(s_{i-1}, s_i) = -\log %\prod\nolimits_{y \in F_i} 
    \alpha_1(I_{F_i})-\log p(s_i|s_{i-1})
\end{align}
\end{statement}
See Supplement \ref{appendix-maximum-likely-state-sequences} for proof. Our reconstruction problem has now become a shortest path problem, and can be solved using one of the several dynamic programming algorithms.

We note that since the path ($s_{1:n}$) defines the subset of the image in the joint probability ($I_{F_{1:n}}$) we can define the probability as a function of only the state sequence
$\bar p(s_{1:n}) \coloneqq  p(s_{1:n}, I_{F_{1:n}}) $
emphasizing that we are solving the most probable path problem.
\subsection{Implementation}

\subsubsection{Fragment generation:}
Fragments are collections of voxels, or ``supervoxels,'' and can be viewed analogously as higher order features such as edgelets or corners.
As described in section \ref{sec:prior}, identifying the subset of fragments that compose the axon, then ordering them becomes equivalent to reconstructing
the axon contour model.

The first step of fragment generation is obtaining a foreground-background mask, which could be obtained, for example, from a neural network, or by simple thresholding. In this work, we use an Ilastik model that was trained on three image subvolumes, each of which has three slices that were labeled \citep{berg2019ilastik}. During prediction, the probability predictions from Ilastik are thresholded at 0.9, a conservative threshold that keeps the number of false positives low.

The connected components of the thresholded image are split into fragments of similar size by identifying the voxel $v$ with the largest predicted foreground probability and placing a ball $B_v$ with radius $7 \, \upmu$m on that voxel. The voxels within $B_v$ are removed and the process is repeated until the component is covered. The component is then split up into pieces by assigning each voxel to the center point from the previous step, $v$, that is closest to it. This procedure ensures that each fragment is no larger than a ball with radius $7 \, \upmu$m. At this size, it is reasonable to assume that each fragment is associated with only one axon branch since no fragment is large enough to extensively cover multiple branches.

Next, the endpoints $x^0,x^1$ and tangents $\tau^0,\tau^1$ are computed as described in Supplement \ref{app:endpoint-comp}.
Each fragment is simplified to the line segment between its endpoints which is rasterized using the Bresenham algorithm \citep{bresenham1965algorithm}. Briefly, the Bresenham algorithm identifies the image axis along which the line segment has the largest range and samples the line once every voxel unit along that axis. Then, the other coordinates are chosen to minimize the distance from the continuous representation line segment.

\subsubsection{Imputing Fragment Deletions}
In practice the imaging data may exhibit significant dropouts leading to significant fragment deletions.
While computing the likelihood of the image data, we augment the gaps between any pair of connected fragments in
$F_1, F_2, \dots $ by augmenting the sequence with imputed fragments
$
F_1, \bar F_1, F_2, \bar F_2, \dots \ .
$, with $\bar F_i \subset D$ the imputed line of voxels which forms the connection between the pair $F_i, F_{i+1}$.
For this define
the start and endpoint of each fragment as $x^0(F) \in \R^3 , x^1 (F) \in \R^3$ with line segment connecting each pair: $$L_{i,i+1} = \{y: y=a x^1(F_i) + (1-a) x^0(F_{i+1}), \ a \in [0,1] \} \ . $$
The imputed fragment $\bar F_{i} \subset D$ for each pair $(F_i, F_{i+1})$ is computed by rasterizing $L_{i,i+1}$ with the Bresenham algorithm.

%$$
%\bar F_{i} =\{ \Delta y_j \in D: \Delta y_j {\displaystyle \cap} L_{i,i+1} \neq \emptyset \}  \ . $$
The likelihood of the sequence of fragments augmented by the imputations becomes
\begin{equation*}
p(s_{1:n},I_{F_{1:n}}) = \prod_{i=2}^n
\alpha_1(I_{F_i}) ^{\delta_{D \setminus F_{1:i-1}}(F_i)}\alpha_1(I_{\bar F_i}) p(s_i| s_{i-1}) p(s_1,I_{F_1})
%\ell(I_{F_{1:n}}| F_{1:n}) 
%p(I_{F_{1:n}}| F_{1:n})
%=
%\left( \prod\nolimits_{i=1}^{n-1} %\ell(I_{F_i};F_i)\ell(I_{\bar F_i}; \bar F_i)
%p(I_{F_i}|F_i)p(I_{\bar F_i}| \bar F_i)
%\right)
%p(I_{F_n}| F_n) \ .
%\ell(I_{F_n}; F_n)
\end{equation*}
\subsubsection{Initial and Endpoint Conditions}
We take the initial conditions to represent $$p(s_1,I_{F_1})= \pi(s_1) p(I_{F_1}|s_1)\ , $$ with $\pi$ the prior on initial state. %with the boundary conditional probability defined as $p(s_{1}|s_0) \coloneqq \pi(s_1)$.
For all of our axon reconstructions we specify an axonal fragment as the start state $s_{start}$ and set $\pi(s_1) \coloneqq \delta_{s_{start}}(s_1)$.

The endpoint conditions are defined
via a user specified terminal state $s_{term}$ where the path ends giving the maximization:
$$\max_{s_{1:n} \in \S^n} p(s_{1:n}, I_{F_{1:n}} 
| s_n=s_{term}) \ .
$$
%This is implemented by modifying the prior to $p(\cdot | s_{term}) = \delta_{s_{term}}(\cdot)$. 
The marginal probability on the terminal state always transitions to itself, so that $p(s_n =s_{term})=\delta_{s_{term}}(s_n)$. Thus, a state sequence solution of length $n$ may end in multiple repetitions of $s_{term}$, 
such as $$s_{1:n}=\{s_1, s_2,...,s_{n'}, s_{term}, s_{term},..., s_{term}\}.$$ 

\subsection{Accuracy Metrics }
\label{accuracy-metrics}
We applied several state of the art reconstruction algorithms to  several neurons in the brain samples from the MouseLight Project from HHMI Janelia \citep{winnubst2019reconstruction}. In this dataset, projection neurons were sparsely labeled then imaged with a two-photon microscope at a voxel resolution of $0.3\times 0.3 \times 1 \upmu$m. Each axon reconstruction is generated semi-automatically by two independent annotators. The MouseLight reconstructions are sampled roughly every $10 \upmu$m, so in some cases we retraced the axons at a higher sampling frequency in order to obtain more precise accuracy metrics.

We quantified reconstruction accuracy using two metrics, the first of which is Frechet distance. Frechet distance is commonly described in the setting of dog walking, where both the dog and owner are following their own predetermined paths. The Frechet distance between the two paths then is the minimum length dog leash needed to complete the walk, where both dog and owner are free to vary their walking speeds but are not allowed to backtrack.  In our setting we compute the Frechet distance between two discrete paths $P:\{1,...,L_p\}\rightarrow \R^3$, $Q:\{1,...,L_q\}\rightarrow \R^3$ as defined in \citet{eiter1994computing}. In this definition, a coupling between $P$ and $Q$ is defined as a sequence of ordered pairs:
\begin{align*}
    (P[a_1],Q[b_1]),(P[a_2],Q[b_2]),...,(P[a_K],Q[b_K])
\end{align*}

where the following conditions are put on $\{a_k\},\{b_k\}$ to ensure that they enumerate through the whole sequences $P$ and $Q$:

\begin{itemize}
    \item $a_1,b_1=1$
    \item $a_N=L_p$, $b_N=L_q$
    \item $a_k=a_{k-1}$ or $a_k=a_{k-1}+1$
    \item $b_k=b_{k-1}$ or $b_k=b_{k-1}+1$
\end{itemize}

Then the discrete Frechet distance is defined as:

\begin{align*}
    \delta_{dF}(P,Q) &= \min_{\text{coupling }\{a_k\},\{b_k\}} \max_{k \in \{1,...,K\}} \left| P\left[a_k\right] - Q\left[b_k\right] \right|
\end{align*}

 We use the standard Euclidean norm for $|\cdot|$. The discrete Frechet distance is an upper bound to the continuous Frechet distance between polygonal curves, and it can be computed more efficiently. Further, if we take a discrete Frechet distance of zero to be an equivalence relation, then $\delta_{dF}$ is a metric on this set of equivalence classes and thus is a natural way to compare non-branching neuronal reconstructions. In this work, all reconstruction are sampled at at least one point per micron.

Various other performance metrics have been proposed, including an arc-length based precision and recall \citep{Wang-Roysam-2011}, a critical node matching based Miss-Extra-Scores (MES) \citep{xie2011anisotropic} and a vertex matching based spatial distance (SD) \citep{peng2010v3d}. We chose to compute SD since it gives a picture of the \textit{average} spatial distance between two reconstructions. This complements the Frechet distance described earlier, which computes the \textit{maximum} spatial distance between two reconstructions.

The first step in computing the SD from reconstruction $P$ to reconstruction $Q$ is, for each point in $P$, finding the distance to the closest point in $Q$. Directed divergence (DDIV) of $P$ from $Q$ is then defined as the average of all these distances. Then, SD is computed by averaging the DDIV from $P$ to $Q$ and the DDIV from $Q$ to $P$.

\section{Acknowledgements}

We thank the MouseLight team at HHMI Janelia for providing us with access to this data, and answering our questions about it.

\section{Author Contributions}

MIM helped to develop the HMM and probabilistic model and DT advised on the theoretical direction of the manuscript. UM coordinated the data acquisition for the experiments. TA designed the study, implemented the software, and managed the manuscript text/figures. All authors contributed to manuscript revision.

\section{Conflict of Interest Statement}

MIM owns a significant share of Anatomy Works with the arrangement being managed by Johns Hopkins
University in accordance with its conflict of interest policies. The
remaining authors declare that the research was conducted in the absence
of any commercial or financial relationships that could be construed as a
potential conflict of interest.

The funders had no role in study design, data collection and analysis,
decision to publish, or preparation of the manuscript.

\section{Funding}
This work is supported by the National Institutes of Health grant RF1MH121539.

\bibliography{main}

\begin{thebibliography}{45}
\providecommand{\natexlab}[1]{#1}
\providecommand{\url}[1]{\texttt{#1}}
\expandafter\ifx\csname urlstyle\endcsname\relax
  \providecommand{\doi}[1]{doi: #1}\else
  \providecommand{\doi}{doi: \begingroup \urlstyle{rm}\Url}\fi

\bibitem[Acciai et~al.(2016)Acciai, Soda, and Iannello]{acciai2016automated}
L.~Acciai, P.~Soda, and G.~Iannello.
\newblock Automated neuron tracing methods: an updated account.
\newblock \emph{Neuroinformatics}, 14\penalty0 (4):\penalty0 353--367, 2016.

\bibitem[Athey et~al.(2021)Athey, Teneggi, Vogelstein, Tward, Mueller, and
  Miller]{athey2021spline}
T.~L. Athey, J.~Teneggi, J.~T. Vogelstein, D.~J. Tward, U.~Mueller, and M.~I.
  Miller.
\newblock Fitting splines to axonal arbors quantifies relationship between
  branch order and geometry.
\newblock \emph{Frontiers in Neuroinformatics}, 2021.

\bibitem[Bellman(1958)]{bellman1958routing}
R.~Bellman.
\newblock On a routing problem.
\newblock \emph{Quarterly of applied mathematics}, 16\penalty0 (1):\penalty0
  87--90, 1958.

\bibitem[Berg et~al.(2019)Berg, Kutra, Kroeger, Straehle, Kausler, Haubold,
  Schiegg, Ales, Beier, Rudy, Eren, Cervantes, Xu, Beuttenmueller, Wolny,
  Zhang, Koethe, Hamprecht, and Kreshuk]{berg2019ilastik}
S.~Berg, D.~Kutra, T.~Kroeger, C.~N. Straehle, B.~X. Kausler, C.~Haubold,
  M.~Schiegg, J.~Ales, T.~Beier, M.~Rudy, K.~Eren, J.~I. Cervantes, B.~Xu,
  F.~Beuttenmueller, A.~Wolny, C.~Zhang, U.~Koethe, F.~A. Hamprecht, and
  A.~Kreshuk.
\newblock ilastik: interactive machine learning for (bio)image analysis.
\newblock \emph{Nature Methods}, Sept. 2019.
\newblock ISSN 1548-7105.
\newblock \doi{10.1038/s41592-019-0582-9}.
\newblock URL \url{https://doi.org/10.1038/s41592-019-0582-9}.

\bibitem[Bresenham(1965)]{bresenham1965algorithm}
J.~E. Bresenham.
\newblock Algorithm for computer control of a digital plotter.
\newblock \emph{IBM Systems journal}, 4\penalty0 (1):\penalty0 25--30, 1965.

\bibitem[Chen et~al.(2015)Chen, Xiao, Liu, and Peng]{chen2015smarttracing}
H.~Chen, H.~Xiao, T.~Liu, and H.~Peng.
\newblock Smarttracing: self-learning-based neuron reconstruction.
\newblock \emph{Brain informatics}, 2\penalty0 (3):\penalty0 135--144, 2015.

\bibitem[Choromanska et~al.(2012)Choromanska, Chang, and
  Yuste]{choromanska2012automatic}
A.~Choromanska, S.-F. Chang, and R.~Yuste.
\newblock Automatic reconstruction of neural morphologies with multi-scale
  tracking.
\newblock \emph{Frontiers in neural circuits}, 6:\penalty0 25, 2012.

\bibitem[Cohen(1991)]{COHEN1991211}
L.~D. Cohen.
\newblock On active contour models and balloons.
\newblock \emph{CVGIP: Image Understanding}, 53\penalty0 (2):\penalty0
  211--218, 1991.
\newblock ISSN 1049-9660.
\newblock \doi{https://doi.org/10.1016/1049-9660(91)90028-N}.
\newblock URL
  \url{https://www.sciencedirect.com/science/article/pii/104996609190028N}.

\bibitem[Cover and Thomas(1991)]{cover1991elements}
T.~M. Cover and J.~A. Thomas.
\newblock \emph{Elements of information theory}, volume~2.
\newblock Wiley, 1991.

\bibitem[Dai et~al.(2019)Dai, Dubois, Arulkumaran, Campbell, Bass, Billot,
  Uslu, De~Paola, Clopath, and Bharath]{dai2019deep}
T.~Dai, M.~Dubois, K.~Arulkumaran, J.~Campbell, C.~Bass, B.~Billot, F.~Uslu,
  V.~De~Paola, C.~Clopath, and A.~A. Bharath.
\newblock Deep reinforcement learning for subpixel neural tracking.
\newblock In \emph{International Conference on Medical Imaging with Deep
  Learning}, pages 130--150. PMLR, 2019.

\bibitem[Dijkstra(1959)]{dijkstra1959note}
E.~W. Dijkstra.
\newblock A note on two problems in connexion with graphs.
\newblock \emph{Numerische mathematik}, 1\penalty0 (1):\penalty0 269--271,
  1959.

\bibitem[Eiter and Mannila(1994)]{eiter1994computing}
T.~Eiter and H.~Mannila.
\newblock Computing discrete fr{\'e}chet distance.
\newblock Technical report, Citeseer, 1994.

\bibitem[Forney(1973)]{forney1973viterbi}
G.~D. Forney.
\newblock The viterbi algorithm.
\newblock \emph{Proceedings of the IEEE}, 61\penalty0 (3):\penalty0 268--278,
  1973.

\bibitem[Friedmann et~al.(2020)Friedmann, Pun, Adams, Lui, Kebschull, Grutzner,
  Castagnola, Tessier-Lavigne, and Luo]{friedmann2020mapping}
D.~Friedmann, A.~Pun, E.~L. Adams, J.~H. Lui, J.~M. Kebschull, S.~M. Grutzner,
  C.~Castagnola, M.~Tessier-Lavigne, and L.~Luo.
\newblock Mapping mesoscale axonal projections in the mouse brain using a 3d
  convolutional network.
\newblock \emph{Proceedings of the National Academy of Sciences}, 117\penalty0
  (20):\penalty0 11068--11075, 2020.

\bibitem[Kass et~al.(1988)Kass, Witkin, and Terzopoulos]{kass1988snakes}
M.~Kass, A.~Witkin, and D.~Terzopoulos.
\newblock Snakes: Active contour models.
\newblock \emph{International journal of computer vision}, 1\penalty0
  (4):\penalty0 321--331, 1988.

\bibitem[Khaneja et~al.(1998)Khaneja, Miller, and
  Grenander]{Khaneja98dynamicprogramming}
N.~Khaneja, M.~I. Miller, and U.~Grenander.
\newblock Dynamic programming generation of curves on brain surfaces, 1998.

\bibitem[Li and Shen(2019)]{li20193d}
Q.~Li and L.~Shen.
\newblock 3d neuron reconstruction in tangled neuronal image with deep
  networks.
\newblock \emph{IEEE transactions on medical imaging}, 39\penalty0
  (2):\penalty0 425--435, 2019.

\bibitem[Li et~al.(2017)Li, Zeng, Peng, and Ji]{li2017deep}
R.~Li, T.~Zeng, H.~Peng, and S.~Ji.
\newblock Deep learning segmentation of optical microscopy images improves 3-d
  neuron reconstruction.
\newblock \emph{IEEE transactions on medical imaging}, 36\penalty0
  (7):\penalty0 1533--1541, 2017.

\bibitem[Li et~al.(2019)Li, Zhu, Li, Bienkowski, Foster, Xu, Ard, Bowman, Zhou,
  Veldman, Yang, Hintiryan, Zhang, and Dong]{Li2019PreciseSO}
R.~Li, M.~Zhu, J.~Li, M.~Bienkowski, N.~N. Foster, H.~Xu, T.~Ard, I.~Bowman,
  C.~Zhou, M.~Veldman, X.~W. Yang, H.~Hintiryan, J.~Zhang, and H.~Dong.
\newblock Precise segmentation of densely interweaving neuron clusters using
  g-cut.
\newblock \emph{Nature Communications}, 10, 2019.

\bibitem[Li et~al.(2020)Li, Quan, Zhou, Huang, Guan, Chen, Xu, Kang, Li, Fu,
  et~al.]{li2020brain}
S.~Li, T.~Quan, H.~Zhou, Q.~Huang, T.~Guan, Y.~Chen, C.~Xu, H.~Kang, A.~Li,
  L.~Fu, et~al.
\newblock Brain-wide shape reconstruction of a traced neuron using the convex
  image segmentation method.
\newblock \emph{Neuroinformatics}, 18\penalty0 (2):\penalty0 199--218, 2020.

\bibitem[Liu et~al.(2018)Liu, Luo, Tan, Wang, and Chen]{liu2018improved}
M.~Liu, H.~Luo, Y.~Tan, X.~Wang, and W.~Chen.
\newblock Improved v-net based image segmentation for 3d neuron reconstruction.
\newblock In \emph{2018 IEEE International Conference on Bioinformatics and
  Biomedicine (BIBM)}, pages 443--448. IEEE, 2018.

\bibitem[napari contributors()]{napari}
napari contributors.
\newblock napari: a multi-dimensional image viewer for python.
\newblock URL \url{doi:10.5281/zenodo.3555620}.

\bibitem[Peng et~al.(2010{\natexlab{a}})Peng, Ruan, Atasoy, and
  Sternson]{peng2010automatic}
H.~Peng, Z.~Ruan, D.~Atasoy, and S.~Sternson.
\newblock Automatic reconstruction of 3d neuron structures using a
  graph-augmented deformable model.
\newblock \emph{Bioinformatics}, 26\penalty0 (12):\penalty0 i38--i46,
  2010{\natexlab{a}}.

\bibitem[Peng et~al.(2010{\natexlab{b}})Peng, Ruan, Long, Simpson, and
  Myers]{peng2010v3d}
H.~Peng, Z.~Ruan, F.~Long, J.~H. Simpson, and E.~W. Myers.
\newblock V3d enables real-time 3d visualization and quantitative analysis of
  large-scale biological image data sets.
\newblock \emph{Nature biotechnology}, 28\penalty0 (4):\penalty0 348--353,
  2010{\natexlab{b}}.

\bibitem[Peng et~al.(2015)Peng, Meijering, and Ascoli]{peng2015diadem}
H.~Peng, E.~Meijering, and G.~A. Ascoli.
\newblock From diadem to bigneuron, 2015.

\bibitem[Peng et~al.(2017)Peng, Zhou, Meijering, Zhao, Ascoli, and
  Hawrylycz]{peng2017automatic}
H.~Peng, Z.~Zhou, E.~Meijering, T.~Zhao, G.~A. Ascoli, and M.~Hawrylycz.
\newblock Automatic tracing of ultra-volumes of neuronal images.
\newblock \emph{Nature methods}, 14\penalty0 (4):\penalty0 332--333, 2017.

\bibitem[Quan et~al.(2016)Quan, Zhou, Li, Li, Li, Li, Lv, Luo, Gong, and
  Zeng]{quan2016neurogps}
T.~Quan, H.~Zhou, J.~Li, S.~Li, A.~Li, Y.~Li, X.~Lv, Q.~Luo, H.~Gong, and
  S.~Zeng.
\newblock Neurogps-tree: automatic reconstruction of large-scale neuronal
  populations with dense neurites.
\newblock \emph{Nature methods}, 13\penalty0 (1):\penalty0 51--54, 2016.

\bibitem[Rabiner and Juang(1986)]{rabiner1986introduction}
L.~Rabiner and B.~Juang.
\newblock An introduction to hidden markov models.
\newblock \emph{ieee assp magazine}, 3\penalty0 (1):\penalty0 4--16, 1986.

\bibitem[Radojevic and Meijering(2019)]{advantra-2019}
M.~Radojevic and E.~Meijering.
\newblock Automated neuron reconstruction from 3d fluorescence microscopy
  images using sequential monte carlo estimation.
\newblock \emph{Neuroinformatics}, 17, 07 2019.
\newblock \doi{10.1007/s12021-018-9407-8}.

\bibitem[Radojević and Meijering(2017)]{radojevic-2017}
M.~Radojević and E.~Meijering.
\newblock {Automated neuron tracing using probability hypothesis density
  filtering}.
\newblock \emph{Bioinformatics}, 33\penalty0 (7):\penalty0 1073--1080, 01 2017.
\newblock ISSN 1367-4803.
\newblock \doi{10.1093/bioinformatics/btw751}.
\newblock URL \url{https://doi.org/10.1093/bioinformatics/btw751}.

\bibitem[Ronneberger et~al.(2015)Ronneberger, Fischer, and
  Brox]{ronneberger2015u}
O.~Ronneberger, P.~Fischer, and T.~Brox.
\newblock U-net: Convolutional networks for biomedical image segmentation.
\newblock In \emph{International Conference on Medical image computing and
  computer-assisted intervention}, pages 234--241. Springer, 2015.

\bibitem[Scott(2015)]{scott2015multivariate}
D.~W. Scott.
\newblock \emph{Multivariate density estimation: theory, practice, and
  visualization}.
\newblock John Wiley \& Sons, 2015.

\bibitem[Snyder and Miller(2012)]{snyder2012random}
D.~L. Snyder and M.~I. Miller.
\newblock \emph{Random point processes in time and space}.
\newblock Springer Science \& Business Media, 2012.

\bibitem[Turetken et~al.(2013)Turetken, Benmansour, Andres, Pfister, and
  Fua]{turetken2013reconstructing}
E.~Turetken, F.~Benmansour, B.~Andres, H.~Pfister, and P.~Fua.
\newblock Reconstructing loopy curvilinear structures using integer
  programming.
\newblock In \emph{Proceedings of the IEEE Conference on Computer Vision and
  Pattern Recognition}, pages 1822--1829, 2013.

\bibitem[Virtanen et~al.(2020)Virtanen, Gommers, Oliphant, Haberland, Reddy,
  Cournapeau, Burovski, Peterson, Weckesser, Bright, {van der Walt}, Brett,
  Wilson, Millman, Mayorov, Nelson, Jones, Kern, Larson, Carey, Polat, Feng,
  Moore, {VanderPlas}, Laxalde, Perktold, Cimrman, Henriksen, Quintero, Harris,
  Archibald, Ribeiro, Pedregosa, {van Mulbregt}, and {SciPy 1.0
  Contributors}]{2020SciPy-NMeth}
P.~Virtanen, R.~Gommers, T.~E. Oliphant, M.~Haberland, T.~Reddy, D.~Cournapeau,
  E.~Burovski, P.~Peterson, W.~Weckesser, J.~Bright, S.~J. {van der Walt},
  M.~Brett, J.~Wilson, K.~J. Millman, N.~Mayorov, A.~R.~J. Nelson, E.~Jones,
  R.~Kern, E.~Larson, C.~J. Carey, {\.I}.~Polat, Y.~Feng, E.~W. Moore,
  J.~{VanderPlas}, D.~Laxalde, J.~Perktold, R.~Cimrman, I.~Henriksen, E.~A.
  Quintero, C.~R. Harris, A.~M. Archibald, A.~H. Ribeiro, F.~Pedregosa, P.~{van
  Mulbregt}, and {SciPy 1.0 Contributors}.
\newblock {{SciPy} 1.0: Fundamental Algorithms for Scientific Computing in
  Python}.
\newblock \emph{Nature Methods}, 17:\penalty0 261--272, 2020.
\newblock \doi{10.1038/s41592-019-0686-2}.

\bibitem[Wang et~al.(2017)Wang, Lee, Pradana, Zhou, and Peng]{wang2017ensemble}
C.-W. Wang, Y.-C. Lee, H.~Pradana, Z.~Zhou, and H.~Peng.
\newblock Ensemble neuron tracer for 3d neuron reconstruction.
\newblock \emph{Neuroinformatics}, 15\penalty0 (2):\penalty0 185--198, 2017.

\bibitem[Wang et~al.(2021)Wang, Zhang, Yu, Song, Liu, Chrzanowski, and
  Cai]{wang2021voxel}
H.~Wang, C.~Zhang, J.~Yu, Y.~Song, S.~Liu, W.~Chrzanowski, and W.~Cai.
\newblock Voxel-wise cross-volume representation learning for 3d neuron
  reconstruction.
\newblock In \emph{International Workshop on Machine Learning in Medical
  Imaging}, pages 248--257. Springer, 2021.

\bibitem[Wang et~al.(2011)Wang, Narayanaswamy, Tsai, and
  Roysam]{Wang-Roysam-2011}
Y.~Wang, A.~Narayanaswamy, C.-L. Tsai, and B.~Roysam.
\newblock A broadly applicable 3-d neuron tracing method based on open-curve
  snake.
\newblock \emph{Neuroinformatics}, 9:\penalty0 193--217, 03 2011.
\newblock \doi{10.1007/s12021-011-9110-5}.

\bibitem[Wang et~al.(2019)Wang, Li, Liu, Zhou, Ruan, Kong, Li, Wang, Zhong,
  Chai, et~al.]{wang2019teravr}
Y.~Wang, Q.~Li, L.~Liu, Z.~Zhou, Z.~Ruan, L.~Kong, Y.~Li, Y.~Wang, N.~Zhong,
  R.~Chai, et~al.
\newblock Teravr empowers precise reconstruction of complete 3-d neuronal
  morphology in the whole brain.
\newblock \emph{Nature communications}, 10\penalty0 (1):\penalty0 1--9, 2019.

\bibitem[Winnubst et~al.(2019)Winnubst, Bas, Ferreira, Wu, Economo, Edson,
  Arthur, Bruns, Rokicki, Schauder, et~al.]{winnubst2019reconstruction}
J.~Winnubst, E.~Bas, T.~A. Ferreira, Z.~Wu, M.~N. Economo, P.~Edson, B.~J.
  Arthur, C.~Bruns, K.~Rokicki, D.~Schauder, et~al.
\newblock Reconstruction of 1,000 projection neurons reveals new cell types and
  organization of long-range connectivity in the mouse brain.
\newblock \emph{Cell}, 179\penalty0 (1):\penalty0 268--281, 2019.

\bibitem[Xiao and Peng(2013)]{APP2}
H.~Xiao and H.~Peng.
\newblock App2: automatic tracing of 3d neuron morphology based on hierarchical
  pruning of a gray-weighted image distance-tree.
\newblock \emph{Bioinformatics}, 29\penalty0 (11), June 2013.

\bibitem[Xie et~al.(2011)Xie, Zhao, Lee, Myers, and Peng]{xie2011anisotropic}
J.~Xie, T.~Zhao, T.~Lee, E.~Myers, and H.~Peng.
\newblock Anisotropic path searching for automatic neuron reconstruction.
\newblock \emph{Medical image analysis}, 15\penalty0 (5):\penalty0 680--689,
  2011.

\bibitem[Yang et~al.(2019)Yang, Hao, Liu, Wan, Zhong, and Peng]{yang2019fmst}
J.~Yang, M.~Hao, X.~Liu, Z.~Wan, N.~Zhong, and H.~Peng.
\newblock Fmst: an automatic neuron tracing method based on fast marching and
  minimum spanning tree.
\newblock \emph{Neuroinformatics}, 17\penalty0 (2):\penalty0 185--196, 2019.

\bibitem[Zhou et~al.(2021)Zhou, Li, Li, Huang, Xiong, Li, Han, Kang, Chen, Li,
  et~al.]{zhou2021gtree}
H.~Zhou, S.~Li, A.~Li, Q.~Huang, F.~Xiong, N.~Li, J.~Han, H.~Kang, Y.~Chen,
  Y.~Li, et~al.
\newblock Gtree: an open-source tool for dense reconstruction of brain-wide
  neuronal population.
\newblock \emph{Neuroinformatics}, 19\penalty0 (2):\penalty0 305--317, 2021.

\bibitem[Zhou et~al.(2018)Zhou, Kuo, Peng, and Long]{zhou2018deepneuron}
Z.~Zhou, H.-C. Kuo, H.~Peng, and F.~Long.
\newblock Deepneuron: an open deep learning toolbox for neuron tracing.
\newblock \emph{Brain informatics}, 5\penalty0 (2):\penalty0 1--9, 2018.

\end{thebibliography}
\bibliographystyle{abbrvnat}

\newpage
%\appendix
\section*{Supplementary Material}
\beginsupplement

\section{Computing Endpoints and Tangents of Fragments}
\label{app:endpoint-comp}

As explained in section \ref{sec:frags}, each fragment is a subset of the image $F \subset D$. Each fragment is assumed to be associated with a segment of an underlying neuron curve, and we want to estimate the locations of the two endpoints of the fragment. First, we compute the length of the diagonal of the bounding box that contains the fragment. We divide this length by two to get $R$. Then, with each voxel $y$ in the fragment, we associate a set of voxels $N_y$ which is the intersection of the voxels in the fragment and the voxels within distance $R$ from $y$. The voxel with the smallest set $N_y$ (by cardinality) is chosen to be the first endpoint. The second endpoint is the voxel that has the smallest set $N_y$ but is also farther than $R$ away from the first endpoint.

Currently the tangents at the endpoints is just approximated by the difference of the endpoints i.e. $\tau^0=\frac{x_0-x_1}{||x_0-x_1||},\tau^1=-\tau^1$. This method is based on the assumption that fragments are small enough that they are approximately straight. We also experimented with approximating the endpoint tangents by computing principal components of voxels near the endpoints, but found it to be less robust for downstream reconstruction.

\section{Curvature Calculation for the Potential in the Markov Chain}
\label{appendix-curvature-approximation}

The term $\kappa(s_{i-1}, s_i)$ is an approximation of the curvature of the path that connects $s_{i-1}$ to $s_i$. For a curve with the tangent vector $T(s)$, curvature is defined as $\kappa(s)=\left|\left| \frac{dT}{ds}\right| \right|$. A finite difference approximation of curvature is then:

\begin{align*}
    \kappa(s)^2 = \left|\left| \frac{dT}{ds}\right| \right|^2 &\approx \left|\left| T(s)-T(s-1) \right| \right|^2 \\
    &= \left|\left| T(s)\right| \right|^2 + \left|\left| T(s-1)\right| \right|^2 - 2 T(s) \cdot T(s-1) \\
    &= 2(1-T(s) \cdot T(s-1)) \\
\end{align*}
Thus,
\begin{align}
    \left|\left| \frac{dT}{ds}\right| \right|^2 &\propto 1-T(s+1) \cdot T(s) \label{eq:curvapprox}
\end{align}

We consider $\tau_i, \tau_{i-1}$ and the normalized vector between the states $\tau_c \coloneqq \frac{x_{i-1}^1 - x_{i}^0}{|x_{i-1}^1 - x_{i}^0|}$ as samples of $T(s)$, then use Eq. \eqref{eq:curvapprox} to estimate the curvature induced by connecting state $s_{i-1}$ to $s_i$:

\begin{align*}
    (\kappa_1)^2 &= 1 - \tau_{i-1}^1 \cdot \tau_c\\
    (\kappa_2)^2 &= 1 - \tau_c \cdot (-\tau_i^0) \\
    \kappa(s_{i-1}, s_i)^2 &= \frac{(\kappa_1)^2+(\kappa_2)^2}{2} &\text{Arithmetic mean}
\end{align*}
\section{Proofs}
\label{appendix-proof-posterior-probability}
\label{appendix-maximum-likely-state-sequences}

Recall the likelihood of a complete fragment under the foreground-background model (Eq. \ref{eq:alpha-product-probability}):
\begin{equation*}
 \alpha_k(I_{F})\coloneqq
\prod_{y \in F} \alpha_k(I_y) \ , k =0,1 \ .
\end{equation*}

{\bf Lemma \ref{Lemma-posterior-probability}:}

For $n > 1$ we have the recursion probability
\begin{subequations}

\begin{align}
    \label{appendix-recursion}
    p(s_{1:n}, I_{D}) = \left(\frac{\alpha_1(I_{F_n})}{\alpha_0(I_{F_n})}\right)^{\delta_{D\setminus F_{1:n-1}}(F_n)} p(s_{n}|s_{n-1} ) p(s_{1:n-1}, I_{D})
\end{align}

\text{implying the factored probability:}

\begin{align}
    p(s_{1:n}, I_{D}) = \prod_{i=2}^n \left(\frac{\alpha_1(I_{F_i})}{\alpha_0(I_{F_i})}\right)^{\delta_{D\setminus F_{1:i-1}}
(F_i)} p(s_{i}|s_{i-1}) p(s_1, I_{D})
\end{align}
\end{subequations}

\begin{proof}
Factor the event $I_D =(I_{F_{1:n}} , I_{D\setminus F_{1:n}})$, then
\begin{eqnarray}
p(s_{1:n}, I_{F_{1:n}} , I_{D\setminus F_{1:n}} )  &=&  p(I_{F_n} | s_{1:n},I_{F_{1:n-1}}, I_{D\setminus F_{1:n} }) p(s_{n}|s_{1:n-1},I_{D\setminus F_{ 1:n}}) p(s_{1:n-1}, I_{F_{1:n-1}}, I_{D\setminus F_{1:n}})
\nonumber \\
&=& p(I_{F_n} | s_n )^{\delta_{D\setminus F_{1:n-1}}(F_n)} p(s_{n}|s_{n-1}) p(s_{1:n-1}, I_{F_{1:n-1}}, I_{D\setminus F_{1:n}})
\nonumber \\
&=& \alpha_1(I_{F_n})^{\delta_{D\setminus F_{1:n-1}}(F_n)} p(s_{n}|s_{n-1}) p(s_{1:n-1}, I_{F_{1:n-1}}, I_{D\setminus F_{1:n}})
\label{factoring-equation}
\end{eqnarray}

We rewrite the last term using the splitting property:

\begin{eqnarray*}
p(s_{1:n-1}, I_{F_{1:n-1}}, I_{D\setminus F_{1:n}})
&=&p (I_{F_{1:n-1}} |s_{1:n-1} , I_{D\setminus F_{1:n}})  p(s_{1:n-1}, I_{D\setminus F_{1:n}})
\\
&=&p (I_{F_{1:n-1}} |s_{1:n-1} )  p( I_{D\setminus F_{1:n}} |s_{1:n-1}) p(s_{1:n-1})
\\
&=&p (I_{F_{1:n-1}} |s_{1:n-1} )  \frac{ p( I_{D\setminus F_{1:n-1}}|s_{1:n-1}) }{ p(I_{F_n}|s_{1:n-1})^{\delta_{D\setminus F_{1:n-1}}(F_n)}
}
 p(s_{1:n-1})
\\
&=&
\frac{1}{\alpha_0(I_{F_n})^{\delta_{D\setminus F_{1:n-1}}(F_n)}}
p(s_{1:n-1}, I_{F_{1:n-1}}, I_{D\setminus F_{1:n-1}})
\end{eqnarray*}

with the last substitution following from the background model.
Substituting into
\ref{factoring-equation} yields the probability written as a recursion \ref{appendix-recursion}:

\begin{equation*}
p(s_{1:n}, I_{D}) =
\left(\frac{\alpha_1(I_{F_n})}{\alpha_0(I_{F_n})}\right)^{\delta_{D\setminus F_{1:n-1}}
(F_n)} p(s_{n}|s_{n-1} )
p(s_{1:n-1}, I_{D})
\end{equation*}
\end{proof}

{\textbf{Statement \ref{maximum-likely-state-sequences}:}}
Define the most probable solution $s_{1:n} \in \S^n$ by the joint probability

$\argmax_{s_{1:n} \in \S^n} p(s_{1:n}, I_{F_{1:n}})$.
Then we have
\begin{equation}
\label{eq:mp_factor}
\max_{s_{1:n} \in \S^n} p(s_{1:n}, I_{F_{1:n}} ) = \max_{s_{1:n} \in \S^n} \prod\nolimits_{i=2}^n  \left( 
%\prod\nolimits_{y \in F_i} 
\alpha_1(I_{F_i})
\right)^{\delta_{D \setminus F_{1:i-1}}(F_i)} p(s_i|s_{i-1}) \,
p(s_1, I_{F_1}) 
\ .
\end{equation}

Further, if $\alpha_1(I_{y})\leq 1$ for all $y$, then the globally optimal solution to the fixed start and end point problem is a nonrepeating state sequence and can be obtained by computing the shortest path in a directed graph where the vertices are the states, and  the edge weight from state $s_{i-1}$ to $s_i$ is given by:

\begin{align}
    e(s_{i-1}, s_i) = -\log %\prod\nolimits_{y \in F_i} 
    \alpha_1(I_{F_i})-\log p(s_i|s_{i-1}) \label{eq:edge_weight}
\end{align}

\begin{proof}

First we will demonstrate the factorization in Eq. \ref{eq:mp_factor}:

\begin{align*}
p(s_{1:n},I_{F_{1:n}}) &= p(s_{1:n},I_{F_{1:n-1}}, I_{F_{n}})
\\ &=  p(I_{F_{n}}| s_{1:n},I_{F_{1:n-1}} )
p(s_{1:n},I_{F_{1:n-1}}) \\
&=p(I_{F_{n}}| s_{1:n},I_{F_{1:n-1}} )
p(I_{F_{1:n-1}}| s_{1:n}) p(s_{1:n})
\\
&=p(I_{F_{n}}| s_{1:n},I_{F_{1:n-1}} )
p(I_{F_{1:n-1}}| s_{1:n-1}) p(s_n|s_{1:n-1}) p(s_{1:n-1})
\\
&=%p(I_{x_{n}}| s_{1:n},I_{x_{1:n-1}} ) 
(\alpha_1(I_{F_n}))^{\delta_{D \setminus F_{1:n-1}}(F_n)}p(s_n|s_{n-1})
p(I_{F_{1:n-1}}| s_{1:n-1})  p(s_{1:n-1})
\ .
\end{align*}

Applying this to factor the conditional probability , $I_{F_{1:i}}$, $i=n-1,\dots,1$ gives the joint product. Next, we will show how this can be solved using shortest path algorithms if $\alpha_1(I_{y})\leq 1$ for all $y$:

First, we want to show that the globally optimal sequence is nonrepeating. This is clear because if $\alpha_1(I_{y})\leq 1$, then every term in the product in equation \eqref{eq:numerator_only} is bounded by 1. Thus, for any state sequence that repeats states, if we remove all elements between the two instances of the repeated states, then this new sequence will have at least as much probability $p(s_{1:n}, I_{F_{1:n}})$.

For nonrepeating state sequences, the probability $p(s_{1:n}, I_{F_{1:n}})$ of \eqref{eq:numerator_only} can be simplified:

\begin{align*}
    p(s_{1:n}, I_{F_{1:n}} ) = \prod\nolimits_{i=2}^n  \left( %\prod\nolimits_{y \in F_i} 
    \alpha_1(I_{F_i}) \right) p(s_i|s_{i-1}) p(s_1, I_{F_1}) \ .
\end{align*}

Taking the logarithm yields a sequentially additive cost function that can be solved with shortest path algorithm on a graph with edge weights given by Equation \eqref{eq:edge_weight}.

\end{proof}

\section{Viterbi Algorithm Counter-example}
\label{viterbi-counter-example}
\begin{figure}[ht]
    \centering
    \includegraphics[width=\textwidth]{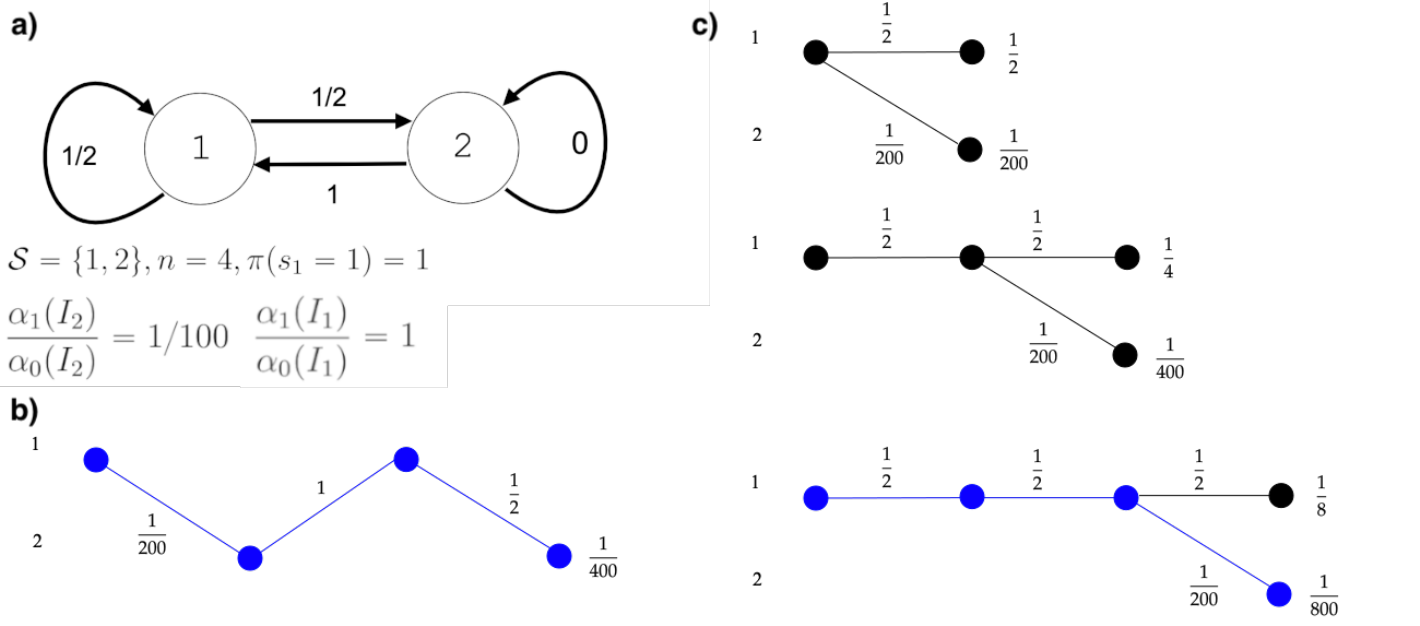}
    \caption{Example showing that the Viterbi algorithm is invalid in our setting. a) Transition probabilities, and image likelihood ratios of a two state Markov model where, in our probabilistic model, the globally optimal state sequence from state $1$ to state $2$ with length $4$ cannot be computed with the classic Viterbi algorithm. Panels b) and c) depict paths through the state space. The nodes represent the states and the line segments represent state transitions, as in Figure 8 of \citet{forney1973viterbi}. Panel b) shows true shortest path from state $1$ to state $2$ with length $4$. Panel c) illustrates three iterations of the Viterbi algorithm, with the lines representing the paths that are stored at each iteration. In this example, the blue path at the bottom of panel c) is identified by naive application of the Viterbi algorithm, but it differs from the true shortest path in panel b). Each transition has a number above it indicating its contribution to the joint probability -  $\left(\frac{\alpha_1(I_{F_i})}{\alpha_0(I_{F_i})}\right)^{\delta_{D\setminus F_{1:i-1}}(F_i)} p(s_{i}|s_{i-1})$ from Eqn. \eqref{eq:joint}. The number at the end of the path is the product of all such terms - the joint probability of the given path.}
    \label{fig:2state}
\end{figure}

Here we present a simple counter example demonstrating that the indicator function in \ref{eq:joint} cannot be ignored, implying the globally optimal solution cannot be obtained efficiently with the Viterbi algorithm.
The details of the state likelihoods and transition probabilities are given in Figure \ref{fig:2state}a.

Here, we remove the indicator function in \ref{eq:joint} and apply the Viterbi algorithm in the usual way. After three iterations, the Viterbi algorithm stores $(1,1,1,2)$ as the highest probability path of length 4 that ends in state 2 (Figure \ref{fig:2state}c). The joint probability of this path is $1/800$. However, there is an alternative path that the algorithm missed, $(1,2,1,2)$, which has higher joint probability $1/400$ (Figure \ref{fig:2state}b). The algorithm did not identify the globally optimal sequence because it did not ``see'' that the cost of transitioning from state 1 to 2 would drop from $1/200$ to $1/2$ after visiting state 2 the first time.

\section{Other Figures}
\label{sec:supp-figs}

\end{document}